%% file: iclr2026_conference.tex
\documentclass{article} % For LaTeX2e
\usepackage{iclr2026_conference,times}

\input{math_commands.tex}

\definecolor{myblue}{rgb}{0.0,0.45,0.8}
\usepackage[colorlinks=true,linkcolor=myblue,citecolor=myblue]{hyperref}
\usepackage{url}
\usepackage{amsfonts}       % blackboard math symbols
\usepackage{amsmath}
\usepackage{amsthm}
\usepackage{amssymb}
\usepackage{array}
\usepackage{nicefrac}       % compact symbols for 1/2, etc.
\usepackage{microtype}      % microtypography
\usepackage{xcolor}         % colors
\usepackage{comment}
\usepackage{booktabs}
\usepackage{multirow} 
\usepackage{graphicx}
\usepackage{enumitem}
\usepackage{pifont}
\usepackage{wrapfig}

\usepackage[most]{tcolorbox}
\usepackage{xcolor}
\usepackage{tikz}

\newcommand*\circled[1]{%
  \tikz[baseline=(char.base)]{
    \node[shape=circle,fill=black,inner sep=0.5pt, minimum size=7pt] (char) 
    {\fontsize{8}{8}\selectfont\textcolor{white}{\textbf{#1}}};
  }%
}

\title{DHG-Bench: A Comprehensive Benchmark for Deep Hypergraph Learning}

\author{Fan Li\textsuperscript{1},\hspace{0.2em}
Xiaoyang Wang\textsuperscript{1},\hspace{0.3em} 
Wenjie Zhang\textsuperscript{1},\hspace{0.2em} 
Ying Zhang\textsuperscript{2},\hspace{0.2em} 
Xuemin Lin\textsuperscript{3}, 
\\
\textsuperscript{1}The University of New South Wales \quad 
\textsuperscript{2}Zhejiang Gongshang University \quad 
\textsuperscript{3}Shanghai Jiaotong University \\
\texttt{\{fan.li8,xiaoyang.wang1,wenjie.zhang\}@unsw.edu.au}\\
\texttt{ying.zhang@zjgsu.edu.cn,} \quad \texttt{xuemin.lin@sjtu.edu.cn}
}

\iclrfinalcopy % Uncomment for camera-ready version, but NOT for submission.
\begin{document}

\maketitle

\begin{abstract}

Deep graph models have achieved great success in network representation learning. However, their focus on pairwise relationships restricts their ability to learn pervasive higher-order interactions in real-world systems, which can be naturally modeled as hypergraphs.
To tackle this issue, Hypergraph Neural Networks (HNNs) have
garnered substantial attention in recent years. 
Despite the proposal of numerous HNNs, the absence of consistent experimental protocols and multi-dimensional empirical analysis impedes deeper understanding and further development of HNN research. While several toolkits for deep hypergraph learning (DHGL) have been introduced to facilitate algorithm evaluation, they provide only limited quantitative evaluation results and insufficient coverage of advanced algorithms, datasets, and benchmark tasks.
To fill the gap, we introduce DHG-Bench, the first comprehensive benchmark for HNNs.
Specifically, DHG-Bench systematically investigates the characteristics of HNNs in terms of four dimensions: effectiveness, efficiency, robustness, and fairness. 
We comprehensively evaluate 17 state-of-the-art HNN algorithms on 22 diverse datasets spanning node-, edge-, and graph-level tasks, under unified experimental settings.
Extensive experiments reveal both the strengths and limitations of existing algorithms, offering valuable insights and directions for future research. 
Furthermore, to facilitate reproducible research, we have developed an easy-to-use library for training and evaluating different HNN methods.
The DHG-Bench library is available at: \url{https://github.com/Coco-Hut/DHG-Bench}.

\begin{comment}
\textit{(i)} insufficient coverage of datasets, algorithms, and tasks; \textit{(ii)} a narrow evaluation of algorithm performance; and \textit{(iii)} inconsistent dataset usage, preprocessing, and experimental setups that hinder comparability.
\end{comment}

\end{abstract}

\section{Introduction}

Graph-structured data has become a ubiquitous tool for modeling the complex relational dependencies among entities in various domains, such as social analysis~\citep{fan2019graph}, e-commerce~\citep{liu2021item}, and finance~\citep{li2024adarisk}. Graph Neural Networks (GNNs) have emerged as the dominant approach for learning on such data, owing to their exceptional ability to leverage both the graph topology and node
attributes. However, many real-world systems involve multi-way or group-wise interactions beyond the pairwise connections of graphs. For instance, multiple authors co-write a paper in co-authorship networks~\citep{yang2022semi}, and groups of proteins interact collectively in biological systems~\citep{kim2024survey}. These higher-order interactions can be naturally modeled by hypergraphs, where each hyperedge connects an arbitrary number of nodes. 
As hypergraphs become increasingly prevalent, there is a growing demand for predictive tasks on them, such as estimating node properties or identifying missing hyperedges~\citep{kim2024survey}.
However, directly applying GNNs to such tasks inevitably collapses higher-order interactions into pairwise relations, resulting in significant information loss and thus sub-optimal performance~\citep{chien2022you}.

To mitigate the aforementioned problem, Hypergraph Neural Networks (HNNs)~\citep{yadati2019hypergcn,chien2022you,wang2023hypergraph,tangtraining} have become the prevailing paradigm for deep hypergraph learning (DHGL), attracting considerable research interest in recent years.
These methods employ neural architectures to transform nodes, hyperedges, and their associated features into vector representations (i.e., embeddings) that effectively preserve higher-order semantics. HNNs have demonstrated state-of-the-art performance across diverse industrial and scientific applications, including product recommendation~\citep{khan2025heterogeneous}, 3D object detection~\citep{fixelle2025hypergraph}, and disease diagnosis~\citep{han2025hypergraph}.

% 思考是否需要在下一段(Despite ...) 中形成问题逻辑：Unfair Comparison --> Incomplete Comparison (insufficient datasets, algorithms, and tasks) ---> Limited Evaluation Dimension (apart from effectiveness)

Despite the emerging studies of HNN algorithms, the comprehensive benchmark for evaluating these
methods remains absent, bringing out the following problems: \textit{(i)} Existing works utilize different
datasets, compared baselines, and experimental setups (e.g., data splitting strategies and parameter settings), which makes it challenging
to achieve a fair comparison. \textit{(ii)} Existing works primarily focus on the effectiveness evaluation of HNN algorithms, while lacking empirical understanding of their efficiency and trustworthiness (e.g., robustness and fairness), both of which are essential for real-world deployment.
This prevents practitioners from understanding the advantages and limitations of HNN algorithms from multiple perspectives and makes it difficult to select appropriate methods for different application scenarios.
Hence, there is an urgent necessity within the community to develop a standardized and comprehensive benchmark for HNNs.

In recent years, several open-sourced toolkits, including HyFER~\citep{hwang2021hyfer}, DHG~\citep{gao2022hgnn+}, and TopoX~\citep{hajij2024topox}, have been proposed to facilitate benchmarkable deep hypergraph learning. 
However, these works provide only limited or even no quantitative performance comparisons, which thus compromises their practical value for practitioners.
Furthermore, they fail to incorporate many state-of-the-art HNN algorithms and provide insufficient coverage of benchmark datasets and evaluation tasks.
Specifically, HyFER supports only the implementation of three HNN models, while the other two libraries include only HNNs proposed before 2023. Moreover, none of these toolkits integrate heterophilic hypergraph datasets, which represent a particularly challenging setting~\citep{li2025hypergraph}, nor do they support graph-level tasks (e.g., hypergraph classification). These limitations significantly restrict the reproducibility and comprehensive evaluation of advanced HNNs.

\begin{comment}
\begin{figure*}[t]
\centering
\includegraphics[width=1\linewidth]{iclr2026/Viz/hnn_timeline.pdf}
\caption{A timeline of the representative hypergraph neural networks}
\label{fig:timeline}
\end{figure*}
\end{comment}

To bridge the gap, we propose DHG-Bench, which serves as the first open-sourced and comprehensive benchmark for HNNs.
Our benchmark encompasses 17 representative HNN methods and 22 diverse hypergraph datasets covering node-level, edge-level, and graph-level tasks.
%We employ standardized computational operators and APIs, consistent data splitting, and processing strategies to ensure a fair comparison.
We employ standardized computational operators and APIs, along with consistent data splitting and processing strategies, to ensure fair comparison.
Beyond effectiveness, our benchmark supports multi-faceted analysis, allowing researchers to investigate the efficiency, robustness, and fairness of current HNN algorithms.
%Based on these four dimensions, we conduct an extensive comparative study of existing HNN algorithms in different scenarios. 
Through extensive experiments, we derive the following key insights: 
\textit{(i)} Existing HNN algorithms exhibit substantial performance variability across datasets and tasks, reflecting their limited generalization ability. %Existing HNN algorithms exhibit substantial performance variability across datasets and tasks, highlighting their limited generalization capacity.
\textit{(ii)} Most HNN methods struggle to strike a satisfactory balance between predictive performance and computational efficiency. %A promising solution lies in decoupling architectures that separate message passing from the training module. 
\textit{(iii)} The performance of HNN algorithms is affected by different types of data perturbations, with feature-level and supervision-level perturbations causing particularly adverse impacts.
%Different types of data perturbations affect the performance of HNN algorithms, with feature-level and supervision-level perturbations having particularly strong impact.
\textit{(iv)} HNN algorithms tend to result in more severe fairness issues than deep models without higher-order message passing, such as MLPs. Our main contributions are summarized as follows:
\begin{itemize}[leftmargin=15pt]
  \item \textbf{Comprehensive Benchmark.} DHG-Bench enables a fair and unified comparison among 17 state-of-the-art HNN methods by standardizing the experimental settings across 22 widely used hypergraph datasets of diverse characteristics. To the best of our knowledge, this is the first comprehensive benchmark for deep hypergraph learning.
  % DHG-Bench enables a fair and unified comparison among 17 state-of-the-art HNN algorithms and 24 diverse hypergraph datasets covering node-level, edge-level, and graph-level tasks. To the best of our knowledge, this is the first comprehensive benchmark for deep hypergraph learning.
  \item \textbf{Multi-dimensional Evaluation and Analysis.} We conduct a systematic analysis of existing HNN methods from various dimensions, encompassing {effectiveness}, {efficiency}, {robustness}, and {fairness}. Extensive experiments uncover the potential strengths and limitations of existing HNN algorithms, offering valuable insights to inform and inspire future research in this field.
  \item \textbf{Open-sourced Benchmark Library.} We release DHG-Bench, an easy-to-use open-sourced benchmark library to support future HNN research. With our toolkit, users can evaluate their algorithms or datasets with less effort.
\end{itemize}

\begin{comment}
For effectiveness, DHG-Bench offers a fair and comprehensive comparison of HNN algorithms on node- and edge-level tasks involving both homophilic and heterophilic hypergraphs. It is also the first benchmark to incorporate the hypergraph-level classification task into evaluation, enabling a more complete assessment of model capabilities across different task levels.
For efficiency, DHG-Bench empirically evaluates the efficiency of representative HNNs in terms of both time and space complexity. 
For robustness, DHG-Bench evaluates HNN models under perturbations at three levels: hypergraph structure, node features, and supervision signals.
For fairness, DHG-Bench evaluates the demographic fairness of existing HNNs through a fair node classification task, revealing their performance disparities across different sensitive groups.
\end{comment}

\section{Preliminary}

% In this section, we first introduce the notations used throughout the paper, and then give the concepts of three benchmark tasks spanning node-, hyperedge-, and hypergraph-level prediction. In this work, we focus on the semi-supervised setting.

Let $\mathcal{G}(\mathcal{V},\mathcal{E},\mathbf{X})$ represent a hypergraph with vertex set $\mathcal{V}$ = $\{v_{i}\}_{i=1}^{|\mathcal{V}|}$ and hyperedge set $\mathcal{E}$ = $\{e_{j}\}_{j=1}^{|\mathcal{E}|}$. $\mathbf{X} \in \mathbb{R}^{|\mathcal{V}| \times F}$ is the node feature matrix with $F$-dimension. 
In this benchmark, we focus on three supervised learning tasks, covering node-, edge-, and graph-level prediction.

\vspace{1mm}
\noindent \textbf{Node Classification.} Given the labeled node set $\mathcal{V}_{L} \subset \mathcal{V}$ with labels $\mathbf{Y}_{L} \in \mathbb{R}^{C}$, where each node $v_{i}$ is associated with a label $y_{i}$ from one of the $C$ classes, the goal of node classification is to train a classifier $f_{\theta}: v \mapsto \mathbb{R}^{C}$ to predict labels $\mathbf{Y}_{U}$ of the remaining unlabeled nodes $\mathcal{V}_{U} = \mathcal{V} \setminus \mathcal{V}_{L}$.

\begin{comment}
Given the labelded node set $\mathcal{V}_{L} \subset \mathcal{V}$ and their 
labels $\mathbf{Y}_{L} \in \mathbb{R}^{C}$, where each node $v_{i}$ is associated with a label $y_{i}$ belonging to one of $C$ classes. Semi-supervised node classification aims to train
a node classifier $f_{\theta}: v \mapsto \mathbb{R}^{C}$ to predict labels $\mathbf{Y}_{U}$ of the remaining nodes $\mathcal{V}_{U} = \mathcal{V} \setminus \mathcal{V}_{L}$.
\end{comment}

\vspace{1mm}
\noindent \textbf{Hyperedge Prediction.} Given a hypergraph $\mathcal{G}(\mathcal{V},\mathcal{E},\mathbf{X})$, we denote $\mathcal{E}^{\prime} \subset 2^{\mathcal{V}} \setminus \mathcal{E}$ as the 
\textbf{target set} which typically consists of (a) unobserved hyperedges or (b) new
hyperedges that will arrive in the near future. Each element in $2^{\mathcal{V}} \setminus \mathcal{E}$ is referred to as a \textbf{hyperedge candidate}, denoted by $c$, as it may belong to $\mathcal{E}^{\prime}$.
The hyperedge prediction task aims to train a hyperedge classifier $f_{\theta}^{\prime}: e \mapsto \{0,1\}$ to predict whether a candidate $c$ belongs to the target set $\mathcal{E}^{\prime}$ or not.

\vspace{1mm}
\noindent \textbf{Hypergraph Classification.} Let $\mathcal{H}$ as the hypergraph set. Given the labeled hypergraph set $\mathcal{H}_{L}$ and 
their labels $\mathbf{Y}_{L} \in \mathbb{R}^{C}$, where each hypergraph $\mathcal{G}_{i}$ is assigned a label $y_{i}$. The hypergraph classification task aims to
train a hypergraph classifier $f_{\theta}^{\prime \prime}: \mathcal{G} \mapsto \mathbb{R}^{C}$ to predict labels $\mathbf{Y}_{U}$ of the unlabeled hypergraphs $\mathcal{H}_{U} = \mathcal{H} \setminus \mathcal{H}_{L}$.

\section{Benchmark Design}
\label{bench_design}

In this section, we introduce the DHG-Bench in terms of datasets (Section~\ref{bench_datasets}), algorithms (Section~\ref{bench_algorithms}), and 
research questions (Section~\ref{bench_questions}) that guide the benchmark study. 
\subsection{Benchmark Datasets}
\label{bench_datasets}

To comprehensively evaluate HNNs, we integrate 22 benchmark datasets
from various domains spanning node-, edge-, and graph-level tasks. In this section, we introduce each dataset category and the corresponding data splitting strategy. Detailed descriptions are provided in Appendix~\ref{datasets_desc}.

\vspace{1mm}
\noindent \textbf{Node-level Classification Datasets.} For the node classification task, we select 13 hypergraph datasets that cover diverse domains and characteristics.
Specifically, we include 8 homophilic datasets: two co-citation networks (Cora and Pubmed~\citep{yadati2019hypergcn}); two co-authorship networks (Cora-CA and DBLP~\citep{yadati2019hypergcn}); two graphics datasets (NTU2012 and ModelNet40~\citep{feng2019hypergraph}); and two hypergraphs that capture user interactions, namely Walmart for co-purchasing~\citep{chien2022you} and Trivago for co-clicking~\citep{kim2023datasets}.
In addition, we consider 5 heterophilic datasets, including two information networks (Actor~\citep{li2025hypergraph} and Yelp~\citep{chien2022you}), an e-commerce network (Amazon-ratings~\citep{li2025hypergraph}), and two social networks (Twitch-gamers and Pokec~\citep{li2025hypergraph}). Moreover, to investigate algorithmic fairness, we include three fairness-sensitive datasets (German, Bail, and Credit~\citep{wei2022augmentations}), which contain sensitive node attributes such as gender, race, and age.
Following~\citep{feng2019hypergraph,chien2022you,tangtraining}, we adopt a split of 50\%/25\%/25\% for training, validation, and testing in the node classification task.

\vspace{1mm}
\noindent \textbf{Hyperedge-level Prediction Datasets.} 
%For the hyperedge prediction task, we adopt four real-world datasets that are widely used in previous studies~\citep{hwang2022ahp,ko2025enhancing}: two co-citation networks (Cora and Pubmed) and two co-authorship networks (Cora-CA and DBLP-CA). To enable a more comprehensive evaluation, we further include two heterophilic datasets, Actor and Pokec, which are characterized by low hyperedge homophily~\citep{li2025hypergraph}.
For the hyperedge prediction task, we use 6 datasets: four widely adopted homophilic academic networks (Cora, Pubmed, Cora-CA, and DBLP-CA)~\citep{hwang2022ahp,ko2025enhancing} and two newly introduced heterophilic datasets, Actor and Pokec~\citep{li2025hypergraph}, which enable a more comprehensive evaluation due to their low hyperedge homophily.
Following~\citep{hwang2022ahp,ko2025enhancing,yu2025hygen}, we randomly split the hyperedges (i.e., positive samples) into training (60\%), validation (20\%), and test (20\%) sets. In addition, we adopt negative sampling (NS)~\citep{yadati2020nhp,hwang2022ahp}, which is devised to enhance the distinguishing ability of the model by introducing non-existing hyperedges as contrastive information for model training.
Specifically, for each training, validation, and test set, we sample an equal number of negative examples as the positive ones. Following~\citep{ko2025enhancing}, we employ a mixed NS strategy that integrates three common heuristic methods, namely sized NS (SNS), motif NS (MNS), and clique NS (CNS)~\citep{patil2020negative}, to increase the diversity of negative samples.

\vspace{1mm}
\noindent \textbf{Hypergraph-level Classification Datasets.} For the hypergraph classification task, we consider 6 benchmark datasets introduced in~\citep{feng2024hypergraph}.
RHG-10 and RHG-3 are two synthetic datasets consisting of distinct high-order structural patterns (e.g., Hyper Pyramid, Hyper Flower, and Hyper Wheel).
IMDB-Dir-Form and IMDB-Dir-Genre are two datasets constructed by the co-director relationship from the original IMDB dataset~\footnote{\url{https://www.imdb.com/}}. Steam-Player is a player-based dataset, where each hypergraph captures tag co-occurrence relationships among games played by a user. 
Twitter-Friend is a social media dataset where each hypergraph represents the friendship network of a specific Twitter user.
For hypergraph classification, following~\citep{feng2024hypergraph}, we adopt an 80\%/10\%/10\% train/validation/test data split.

\subsection{Benchmark Algorithms}
\label{bench_algorithms}

We integrate 17 state-of-the-art HNN algorithms across three mainstream categories: spectral-based, spatial-based, and tensor-based methods. In addition, we include MLP and two GNN-based methods, CEGCN and CEGAT~\citep{chien2022you}, as baselines. Detailed descriptions are provided in Appendix~\ref{algo_details}.
We rigorously reproduce all methods according to their papers and source codes.

% To ensure a fair evaluation, we perform hyperparameter tuning with a reasonable search budget on all datasets for HNN methods. %Details of the algorithmic settings in our experiments are provided in Appendix~\ref{exp_details}.

%We have integrated 16 representative and competitive HNN algorithms in DHG-Bench. They are divided into three technique categories: Spectral-based HNN, Spatial-based HNN, and Tensor-based HNN. We briefly introduce each category in the following, and more details are provided in Appendix~\ref{algo_details}.

\vspace{1mm}
\noindent \textbf{Spectral-based HNNs.} 
Spectral-based HNNs perform message propagation and feature transformation by applying spectral convolution defined through Laplacian operators of hypergraphs~\citep{wang2024t}.
%Spectral-based HNNs leverage the spectral theory of hypergraph Laplacians for message propagation and feature transformation. They define convolution in the spectral domain via eigendecomposition of the hypergraph Laplacian derived from the adjacency matrix~\citep{saxena2024dphgnn,wang2024t}.
We implement 10 representative algorithms including HGNN~\citep{feng2019hypergraph}, HyperGCN~\citep{yadati2019hypergcn}, HCHA~\citep{bai2021hypergraph}, LEGCN~\citep{yang2022semi}, HyperND~\citep{prokopchik2022nonlinear}, PhenomNN~\citep{wang2023hypergraph},
SheafHyperGNN~\citep{duta2023sheaf}, HJRL~\citep{yan2024hypergraph}, DPHGNN~\citep{saxena2024dphgnn}, and TF-HNN~\citep{tangtraining}.

\vspace{1mm}
\noindent \textbf{Spatial-based HNNs.} 
\begin{comment}
In contrast to spectral-based method, spatial-based methods focus on the local
connectivity of each node without going to the spectral domain~\citep{wang2024t}. 
They commonly apply a two-stage neighborhood aggregation scheme to learn hypergraph representations, which entails updating the hyperedge embedding by aggregating embeddings of its incident
nodes, and updating the node representation by propagating
information from representations of its incident hyperedges~\citep{gao2022hgnn+,wang2024t}. 
\end{comment}
Unlike spectral methods, spatial-based HNNs focus on local connectivity without entering the spectral domain, typically learning representations through two-stage neighborhood aggregation: updating hyperedges from incident nodes and updating nodes from incident hyperedges.
We incorporate 5 typical algorithms including HNHN~\citep{dong2020hnhn}, UniGNN~\citep{huang2021unignn}, AllSetTransformer~\citep{chien2022you}, ED-HNN~\citep{wang2023equivariant}, and HyperGT~\citep{liu2024hypergraph}. 
For UniGNN with multiple variants (e.g., UniGAT, UniGIN, and UniGCNII), we report only UniGCNII, the most competitive variant identified in the original paper, while our open-sourced library also supports the implementations of other variants.

\vspace{1mm}
\noindent \textbf{Tensor-based HNNs.} Tensor-based methods leverage tensor operations that provide a structured and effective means
of capturing the complexity of hypergraph interactions~\citep{wang2025generalization}. We select two representative algorithms: EHNN~\citep{kim2022equivariant} and T-HyperGNN~\citep{wang2024t}.

\subsection{Research Questions}
\label{bench_questions}

We systematically design the DHG-Bench to comprehensively evaluate the existing HNN algorithms and
inspire future research. In particular, we aim to investigate the following research questions.

\vspace{1mm}

\noindent \textbf{RQ1: How much progress has been made by existing HNN methods?} 

\vspace{1mm}

\noindent \textbf{Motivation and Experiment Design.} Previous research on HNNs has been limited by inconsistent experimental settings and insufficient coverage of datasets, algorithms, and tasks, thereby hindering fair and comprehensive evaluation of different methods.
Given the standardized experimental environment provided by DHG-Bench, 
the first question is to revisit the progress of existing HNN methods and identify potential directions for enhancement. A high-quality HNN method is expected to perform consistently well across different datasets and application scenarios.
To answer this question, we evaluate the performance of HNN methods on diverse, widely used hypergraph datasets across three benchmark tasks: node classification, hyperedge prediction, and hypergraph classification. Detailed experimental settings can be found in Appendix~\ref{exp_setups}. % Evaluation results are presented in Section~\ref{exp:effectiveness}.

\begin{comment}
Previous research on HNNs has been hindered by inconsistent data preprocessing, varying hyperparameter settings, and limited benchmark tasks, making it difficult to fairly and comprehensively evaluate the performance of different methods.
Given the standardized experimental environment provided by DHG-Bench, investigating the effectiveness of various HNN algorithms aims to deepen our understanding of their strengths and limitations across diverse tasks and domain data, and to identify avenues for potential improvement.
To answer this question, we evaluate the performance of HNN methods on 21 widely used hypergraph datasets across three benchmark tasks: node classification, hyperedge prediction, and hypergraph classification. Detailed experimental settings can be found in Appendix~\ref{exp_details}.
\end{comment}

\vspace{1mm}

\noindent \textbf{RQ2: How efficient are these HNN methods in terms of time and space?} 

\vspace{1mm}

\noindent \textbf{Motivation and Experiment Design.} 
Training the message-passing module of HNNs makes loss computation
interdependent for connected nodes, resulting in intensive computational demands and substantial memory constraints.
However, the efficiency and scalability of HNN algorithms have been largely overlooked. 
A thorough understanding of the trade-off between computational cost and predictive performance is essential for assessing their suitability for real-time and large-scale applications.
%despite being critical for understanding the trade-off between computational cost and predictive performance. A thorough evaluation of these aspects is essential for assessing the practical suitability of each method for real-time and large-scale deployment.
To answer this question, we perform node classification, the most widely used benchmark task, on datasets of varying scales (Cora, DBLP-CA, Yelp, and Trivago), reporting the training time to reach the best validation performance and the peak GPU memory consumption.

\vspace{1mm}

\noindent \textbf{RQ3: Are existing HNN methods robust to different types of data perturbations?} 

\vspace{1mm}

\noindent \textbf{Motivation and Experiment Design.} 
Real-world hypergraph data inevitably contains noise, task-irrelevant information, or even mistakes~\citep{cai2022hypergraph}. 
A reliable HNN should maintain stable performance when exposed to such noisy data, particularly in high-stakes domains such as healthcare and finance~\citep{caihypernear}, where inaccurate decisions can adversely affect individual lives or broader societal systems.
Evaluating the robustness of HNNs not only reveals potential vulnerabilities in existing methods but also guides the development of more resilient models.
To answer this question, we simulate realistic data perturbations from three perspectives: structure, feature, and supervision signals. 
For each perturbation type, we vary the noise intensity and subsequently train and test HNNs on the corresponding modified hypergraph.
Detailed experimental settings are in Appendix~\ref{robust_setups}.

\vspace{1mm}

\noindent \textbf{RQ4: Do existing HNN methods yield unbiased predictions across demographic groups?} 

\vspace{1mm}

\noindent \textbf{Motivation and Experiment Design.} Fairness has recently emerged as a critical concern in graph machine learning (GML)~\citep{dong2023fairness}. Prior studies have shown that representations learned by GNNs can result in biased predictions, often favoring certain demographic groups defined by sensitive attributes (e.g., gender and race)~\citep{ling2023learning,zhu2024fair,yang2024fairsin}. Such bias hinders the deployment of GML models in high-stakes applications such as crime prediction~\citep{suresh2019framework} and credit evaluation~\citep{yeh2009comparisons}.
Despite its importance, fairness in deep hypergraph learning has received little attention. To the best of our knowledge, this work presents the first benchmark evaluation of fairness in this context, which is crucial for developing ethically sound and trustworthy HNN models.
To answer this question, we conduct node classification on three fairness-sensitive datasets (German, Bail, and Credit~\citep{wei2022augmentations}), each of which contains demographic-sensitive attributes. We assess algorithmic fairness using two widely adopted group fairness metrics: demographic parity ($\Delta_{DP}$)~\citep{dwork2012fairness}, and equalized odds ($\Delta_{EO}$)~\citep{hardt2016equality}. 
The detailed descriptions of the two metrics can be found in Appendix~\ref{eval_metrics}.

\section{Experiment Results and Analysis}

\subsection{Effectiveness Evaluation (\textbf{\textcolor{magenta}{RQ1}})}
\label{exp:effectiveness}

\begin{table*}[t]
    \centering
    \small
    \caption{Evaluation results of node classification: mean accuracy (\%) ± standard deviation. The best results are shown in \textbf{\textcolor{magenta}{bold}} and the runner-ups are \underline{underlined}. OOM denotes the out-of-memory issue.} %  (the same for tables below).
    \resizebox{\linewidth}{!}{
    \begin{tabular}{c c c c c c c c c c c}
        \toprule
        \textbf{Method} & \textbf{Cora} & \textbf{Pubmed} & \textbf{Cora-CA} & \textbf{DBLP-CA} & \textbf{Walmart} & \textbf{Trivago} & \textbf{Actor} & \textbf{Gamers} & \textbf{Pokec} & \textbf{Yelp} \\
        \midrule
        MLP & 75.33\scriptsize{$\pm$0.88} & 86.62\scriptsize{$\pm$0.26} & 75.57\scriptsize{$\pm$1.08} & 85.54\scriptsize{$\pm$0.15} & 63.21\scriptsize{$\pm$0.12} & 36.76\scriptsize{$\pm$0.66} & \underline{86.06\scriptsize{$\pm$0.36}} & \textbf{\textcolor{magenta}{52.57\scriptsize{$\pm$0.49}}} & \underline{59.64\scriptsize{$\pm$0.48}} & 31.84\scriptsize{$\pm$0.45} \\
        CEGCN & 76.90\scriptsize{$\pm$0.75} & 86.03\scriptsize{$\pm$0.39} & 78.40\scriptsize{$\pm$1.25} & 89.75\scriptsize{$\pm$0.33} & 70.40\scriptsize{$\pm$0.18} & 47.24\scriptsize{$\pm$1.09} & 67.41\scriptsize{$\pm$0.29} & 51.02\scriptsize{$\pm$0.53} & 57.37\scriptsize{$\pm$0.38} & OOM \\
        CEGAT & 77.22\scriptsize{$\pm$1.03} & 86.09\scriptsize{$\pm$0.51} & 78.02\scriptsize{$\pm$1.24} & 89.61\scriptsize{$\pm$0.22} & 65.83\scriptsize{$\pm$0.92} & OOM & 73.87\scriptsize{$\pm$0.83} & 51.05\scriptsize{$\pm$0.61} & 57.34\scriptsize{$\pm$0.52} & OOM \\
        \midrule
        HGNN & 77.90\scriptsize{$\pm$1.17} & 86.17\scriptsize{$\pm$0.52} & 82.84\scriptsize{$\pm$0.46} & 91.00\scriptsize{$\pm$0.27} & 77.12\scriptsize{$\pm$0.12} & 57.67\scriptsize{$\pm$1.61} & 77.83\scriptsize{$\pm$0.37} & 52.38\scriptsize{$\pm$0.56} & 57.87\scriptsize{$\pm$0.76} & 33.71\scriptsize{$\pm$0.24} \\
        HyperGCN & 78.38\scriptsize{$\pm$1.63} & 87.42\scriptsize{$\pm$0.42} & 81.65\scriptsize{$\pm$1.58} & 89.51\scriptsize{$\pm$0.18} & 68.75\scriptsize{$\pm$0.56} & 42.39\scriptsize{$\pm$1.25} & 81.82\scriptsize{$\pm$0.39} & 51.32\scriptsize{$\pm$0.72} & 57.51\scriptsize{$\pm$0.54} & 29.29\scriptsize{$\pm$0.55} \\
        HCHA & 77.84\scriptsize{$\pm$1.23} & 86.33\scriptsize{$\pm$0.54} & 83.01\scriptsize{$\pm$0.58} & 91.18\scriptsize{$\pm$0.30} & 77.66\scriptsize{$\pm$0.18} & 52.50\scriptsize{$\pm$3.43} & 78.30\scriptsize{$\pm$0.47} & 52.35\scriptsize{$\pm$0.71} & 58.19\scriptsize{$\pm$0.45} & 33.13\scriptsize{$\pm$0.23} \\
        LEGCN & 74.36\scriptsize{$\pm$1.03} & 87.52\scriptsize{$\pm$0.50} & 74.59\scriptsize{$\pm$1.04} & 85.16\scriptsize{$\pm$0.14} & 62.98\scriptsize{$\pm$0.09} & 33.45\scriptsize{$\pm$1.45} & 85.34\scriptsize{$\pm$0.45} & 51.31\scriptsize{$\pm$0.65} & \textbf{\textcolor{magenta}{59.66\scriptsize{$\pm$0.63}}} & OOM \\
        HyperND & 79.23\scriptsize{$\pm$0.63} & 86.73\scriptsize{$\pm$0.56} & 83.19\scriptsize{$\pm$0.71} & 91.34\scriptsize{$\pm$0.19} & 75.10\scriptsize{$\pm$0.54} & \underline{87.19\scriptsize{$\pm$1.89}} & 83.19\scriptsize{$\pm$0.92} & 52.39\scriptsize{$\pm$0.60} & 57.65\scriptsize{$\pm$1.08} & OOM \\
        PhenomNN & 78.97\scriptsize{$\pm$1.41} & 87.81\scriptsize{$\pm$0.12} & 84.05\scriptsize{$\pm$1.05} & \textbf{\textcolor{magenta}{91.83\scriptsize{$\pm$0.25}}} & OOM & OOM & 83.14\scriptsize{$\pm$0.49} & 51.80\scriptsize{$\pm$0.73} & 58.43\scriptsize{$\pm$0.92} & OOM \\
        SheafHyperGNN & 79.03\scriptsize{$\pm$0.90} & 87.10\scriptsize{$\pm$0.47} & \underline{84.08\scriptsize{$\pm$0.50}} & 91.09\scriptsize{$\pm$0.31} & OOM & OOM & 85.00\scriptsize{$\pm$0.32} & 52.07\scriptsize{$\pm$0.53} & 59.06\scriptsize{$\pm$0.37} & OOM \\
        HJRL & 78.67\scriptsize{$\pm$1.47} & \textbf{\textcolor{magenta}{87.98\scriptsize{$\pm$0.49}}} & 83.72\scriptsize{$\pm$0.74} & OOM & OOM & OOM & 71.54\scriptsize{$\pm$0.64} & 51.62\scriptsize{$\pm$0.61} & 57.57\scriptsize{$\pm$0.47} & OOM \\
        DPHGNN & 76.40\scriptsize{$\pm$1.36} & 86.72\scriptsize{$\pm$0.33} & 82.13\scriptsize{$\pm$1.13} & OOM & OOM & OOM & 83.65\scriptsize{$\pm$0.59} & 52.36\scriptsize{$\pm$0.59} & 58.20\scriptsize{$\pm$0.58} & OOM \\
        TF-HNN & \textbf{\textcolor{magenta}{79.47\scriptsize{$\pm$1.31}}} & \underline{87.90\scriptsize{$\pm$0.37}} & \textbf{\textcolor{magenta}{84.19\scriptsize{$\pm$0.89}}} & 91.38\scriptsize{$\pm$0.24} & 77.04\scriptsize{$\pm$0.12} & \textbf{\textcolor{magenta}{90.79\scriptsize{$\pm$0.79}}} & 85.96\scriptsize{$\pm$0.41} & 52.34\scriptsize{$\pm$0.53} & 59.17\scriptsize{$\pm$0.52} & \textbf{\textcolor{magenta}{35.16\scriptsize{$\pm$0.54}}} \\
        \midrule
        HNHN & 75.24\scriptsize{$\pm$1.38} & 85.66\scriptsize{$\pm$1.28} & 76.51\scriptsize{$\pm$1.34} & 85.84\scriptsize{$\pm$0.07} & 65.21\scriptsize{$\pm$0.28} & 53.75\scriptsize{$\pm$1.43} & 81.20\scriptsize{$\pm$0.36} & 51.12\scriptsize{$\pm$0.65} & 58.55\scriptsize{$\pm$0.93} & 25.86\scriptsize{$\pm$0.63} \\
        UniGNN & \underline{79.41\scriptsize{$\pm$1.24}} & 87.57\scriptsize{$\pm$0.54} & 83.49\scriptsize{$\pm$1.58} & \underline{91.71\scriptsize{$\pm$0.20}} & 76.26\scriptsize{$\pm$0.58} & 36.15\scriptsize{$\pm$0.56} & 84.61\scriptsize{$\pm$0.46} & \underline{52.50\scriptsize{$\pm$0.57}} & 58.56\scriptsize{$\pm$0.73} & 31.09\scriptsize{$\pm$0.61} \\
        AllSetTransformer & 78.02\scriptsize{$\pm$1.43} & 87.79\scriptsize{$\pm$0.30} & 82.95\scriptsize{$\pm$0.62} & 91.51\scriptsize{$\pm$0.22} & \textbf{\textcolor{magenta}{78.61\scriptsize{$\pm$0.13}}} & 59.92\scriptsize{$\pm$4.02} & 85.66\scriptsize{$\pm$0.41} & 51.74\scriptsize{$\pm$0.75} & 58.55\scriptsize{$\pm$0.56} & 33.18\scriptsize{$\pm$0.88} \\
        ED-HNN & 78.58\scriptsize{$\pm$0.52} & 87.65\scriptsize{$\pm$0.23} & 82.98\scriptsize{$\pm$0.93} & 91.55\scriptsize{$\pm$0.19} & 77.90\scriptsize{$\pm$0.21} & 75.99\scriptsize{$\pm$2.60} & 85.77\scriptsize{$\pm$0.46} & 50.54\scriptsize{$\pm$0.23} & 58.68\scriptsize{$\pm$0.40} & \underline{34.84\scriptsize{$\pm$0.93}} \\
        HyperGT & 75.57\scriptsize{$\pm$1.11} & 86.06\scriptsize{$\pm$0.54} & 75.42\scriptsize{$\pm$0.62} & 84.53\scriptsize{$\pm$0.30} & OOM & OOM & 84.43\scriptsize{$\pm$0.47} & 51.19\scriptsize{$\pm$0.57} & 57.73\scriptsize{$\pm$0.76} & OOM \\
        \midrule
        EHNN & 76.51\scriptsize{$\pm$1.52} & 87.12\scriptsize{$\pm$0.31} & 81.68\scriptsize{$\pm$0.81} & 90.47\scriptsize{$\pm$0.43} & \underline{77.95\scriptsize{$\pm$0.14}} & OOM & \textbf{\textcolor{magenta}{86.21\scriptsize{$\pm$0.49}}} & 52.14\scriptsize{$\pm$0.76} & 58.23\scriptsize{$\pm$1.07} & 34.09\scriptsize{$\pm$3.19} \\
        T-HyperGNN & 74.20\scriptsize{$\pm$1.37} & 86.28\scriptsize{$\pm$0.62} & 75.01\scriptsize{$\pm$1.44} & 85.44\scriptsize{$\pm$0.14} & 73.48\scriptsize{$\pm$0.33} & OOM & 85.32\scriptsize{$\pm$0.48} & 51.82\scriptsize{$\pm$0.38} & 58.82\scriptsize{$\pm$0.49} & OOM \\
        \bottomrule
    \end{tabular}}
    \label{tab:node_cls}
\end{table*}

To investigate the effectiveness of existing HNNs, we compare their performance across benchmark tasks at the node, edge, and graph levels. Due to space constraints, additional node classification results on NTU2012, ModelNet40, and Ratings (Table~\ref{tab:add_node_cls}), as well as the complete results of hyperedge prediction (Table~\ref{tab:edge_pred}) and hypergraph classification (Table~\ref{tab:graph_cls}), are available in Appendix~\ref{exp:add_effect}.
%To investigate the effectiveness of existing HNNs, we compare their performance across benchmark tasks at the node, edge, and graph levels. Due to space constraints, additional results of node classification on NTU2012, ModelNet40, and Ratings (Table~\ref{tab:add_node_cls}), and full experimental results of hyperedge prediction (Table~\ref{tab:edge_pred}) and hypergraph classification (Table~\ref{tab:graph_cls}) are available in Appendix~\ref{exp:add_effect}.

\subsubsection{Effectiveness on Node Classification Task}
\noindent \textbf{Results} (Table~\ref{tab:node_cls} and Table~\ref{tab:add_node_cls}). 
\circled{1} Across diverse datasets, HNNs generally outperform both CEGCN and CEGAT, suggesting that naively extending GNNs to hypergraphs via clique expansion disrupts high-order structures and degrades predictive performance.
This highlights the necessity of designing neural architectures with dedicated high-order message passing.
\circled{2} HNNs achieve notable improvements over MLP on homophilic datasets, but on heterophilic datasets, most HNNs even underperform MLP, which only leverages node features.
This reveals the adverse impact of heterophilic connections on hypergraph representation learning and underscores the need to rethink HNN design in such settings.
\circled{3} TF-HNN consistently ranks among the top-performing methods across diverse datasets, achieving optimal or near-optimal results. Moreover, unlike other advanced HNNs (e.g., PheomNN, DPHGNN, and HyperGT) that fail on large-scale datasets due to out-of-memory issues, TF-HNN remains scalable.
These findings underscore the promise of its decoupled architecture for enhanced generalization and scalability.
%These findings highlight the potential of training-free message passing and a decoupled architecture to enable efficient hypergraph learning with strong generalization.

\subsubsection{Effectiveness on Hyperedge Prediction Task}
\noindent \textbf{Results} (Table~\ref{tab:edge_pred}). 
\circled{1} Advanced HNN methods that generally achieve superior performance on node classification fail to maintain the same level of competitiveness in hyperedge prediction.
Specifically, the two earliest methods, HGNN and HyperGCN, along with the tensor-based EHNN introduced in 2022, collectively achieve all the best results and the majority of second-best results across the six hyperedge prediction datasets. In contrast, recent HNNs (e.g., ED-HNN, HJRL, DPHGNN, TF-HNN) often show a notable performance gap compared to the above three. 
For example, on DBLP-CA, TF-HNN achieves an AUROC of 75.70\% and an AP of 74.97\%, which are 13.76\% and 16.70\% lower than those of the best-performing model, HyperGCN.
\circled{2} Across hyperedge prediction benchmarks, HNN algorithms display considerable performance divergence depending on the dataset, and none consistently deliver the best results.
For instance, while EHNN achieves state-of-the-art performance on Cora and Pubmed, it obtains only 77.83\% AUROC on Cora-CA, ranking 11th among 17 HNNs and 14.90\% lower than the top-performing HyperGCN.

\subsubsection{Effectiveness on Hypergraph Classification Task}
\noindent \textbf{Results} (Table~\ref{tab:graph_cls}).
\circled{1} HNN algorithms perform markedly better on synthetic datasets than on real-world ones. On RHG-10, most models achieve over 90\% accuracy and Macro-F1, and on RHG-3, many even exceed 98\%. In contrast, on real-world datasets, accuracies rarely surpass 70\%, reflecting the structural complexity of real hypergraphs. This gap underscores the need for more realistic and challenging benchmarks to rigorously evaluate hypergraph classification.
\circled{2} HNN methods generally outperform GNN-based approaches built on clique expansion, as the latter often distorts global hypergraph structures, whereas higher-order message passing in HNNs preserves these dependencies and enhances discriminative power.
\circled{3} HNNs' performance varies considerably across datasets, with no method demonstrating consistent superiority. For instance, while DPHGNN achieves the best accuracy on IMDB-Dir-Form, it falls to 11th on IMDB-Dir-Genre and 14th on Steam-Player across all evaluated HNNs, underscoring the substantial impact of dataset characteristics on model performance.
\circled{4} Many HNN methods fail to achieve a desirable trade-off between accuracy and Macro-F1. For example, on the Twitter dataset, HNHN achieves 58.47\% accuracy (third highest among all HNN models) but only 39.40\% Macro-F1, the lowest overall. 

\begin{tcolorbox}[colback=white, colframe=black, sharp corners, boxrule=0.5pt,
  left=2pt, right=2pt, top=2pt, bottom=2pt, fonttitle=\normalsize, breakable]
\textbf{\textcolor{magenta}{Key Insights for RQ1:}} 
HNN algorithms display varying levels of effectiveness across predictive tasks. While advanced HNNs achieve strong results on node-level tasks, they often fail to deliver superior performance on edge- and graph-level tasks. 
Moreover, the predictive capability of HNNs is highly sensitive to dataset characteristics, with data heterophily substantially impairing learning on hypergraphs. These findings highlight the need for future research to enhance the generalization and adaptability of hypergraph models across diverse tasks and datasets.
\end{tcolorbox}

\subsection{Efficiency and Scalability Evaluation (\textbf{\textcolor{magenta}{RQ2}})}
\label{exp:efficiency}

\begin{figure*}[t]
\centering
\includegraphics[width=1\linewidth]{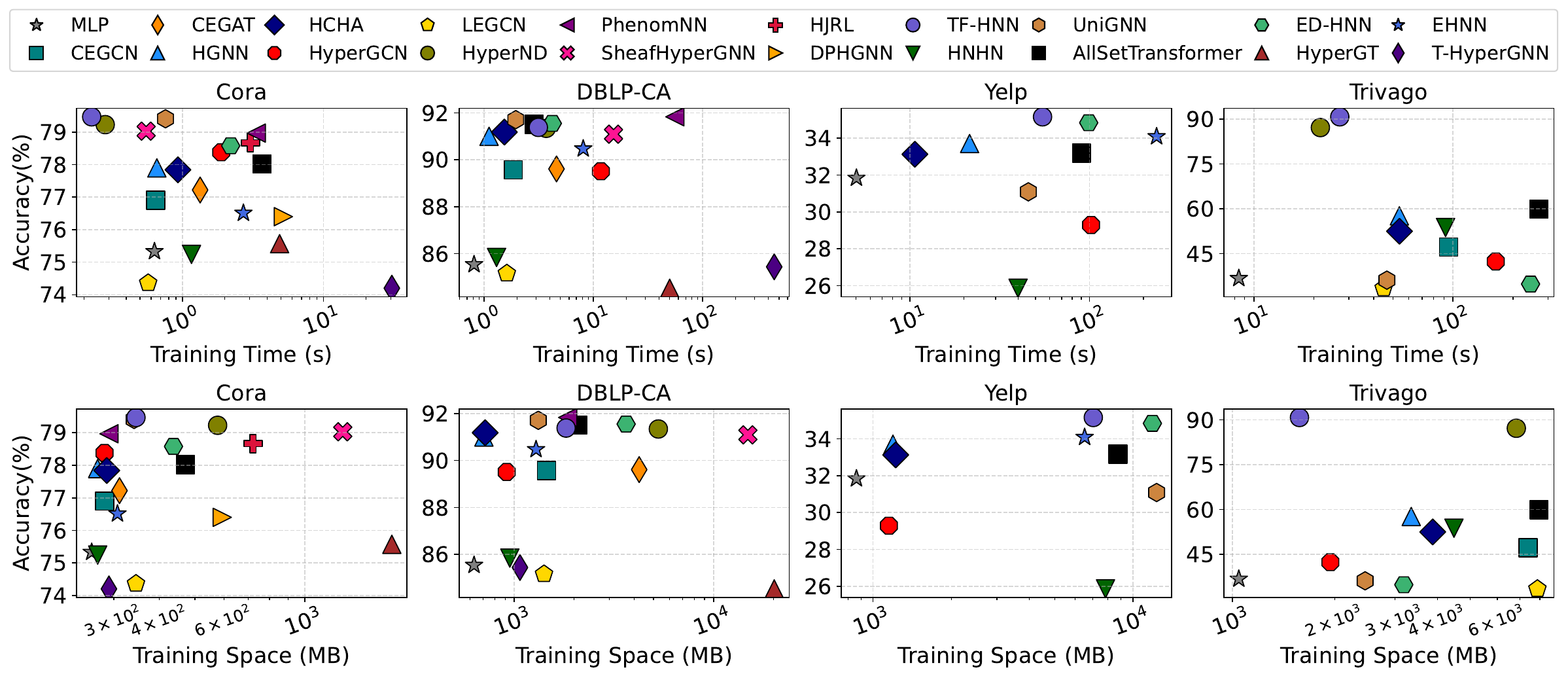}
\caption{Training time and space analysis on Cora, DBLP-CA, Yelp, and Trivago.}
\label{fig:efficiency}
\end{figure*}

%In this section, we evaluate the training time and memory consumption of HNNs on datasets of varying scales (Cora, DBLP-CA, Yelp, and Trivago). The results are presented in Figure~\ref{fig:efficiency}. 

% 可以加一个结论说明和MLP的比较，说明message-passing范式的耗时

\noindent \textbf{Results} (Figure~\ref{fig:efficiency}). 
\circled{1} CEGCN and CEGAT face scalability challenges on large datasets (e.g., Yelp and Trivago), where clique expansion produces dense edges and leads to significant training memory overhead.
\circled{2} Most advanced HNN methods struggle to achieve a satisfactory balance between model utility and efficiency.
For example, on the Yelp dataset, ED-HNN and EHNN provide only marginal accuracy gains over the simple HGNN, yet their training times are over 9× and 23× longer, respectively, reflecting a substantial rise in computational cost.
In addition, many HNNs suffer from memory bottlenecks on large-scale datasets. Specifically, on Yelp, 8 out of 17 methods encounter out-of-memory (OOM) issues. On Trivago, although 10 HNNs remain computationally scalable, most fail to deliver satisfactory predictive performance. Only TF-HNN (90.79\%) and HyperND (87.19\%) achieve accuracy above 60\%. This may result from the intricate patterns of large-scale graphs.
\circled{3} Tensor-based approaches exhibit more pronounced efficiency and scalability limitations than the other two kinds of methods. 
T-HyperGNN can only scale to the medium-sized DBLP-CA dataset, where it runs approximately 406 times slower than the fastest method, HGNN. Moreover, on Yelp, EHNN incurs the longest training time and fails to scale to the large-scale Trivago dataset.
\circled{4} Among all evaluated methods, TF-HNN generally achieves a superior trade-off between utility and both time and space efficiency.
For example, on the large-scale Trivago dataset, it achieves the best predictive performance with no more than 1.6 GB of memory and under 30 seconds of runtime, ranking first in memory efficiency and second in training time among all HNN methods.

\begin{tcolorbox}[colback=white, colframe=black, sharp corners, boxrule=0.5pt,
  left=2pt, right=2pt, top=2pt, bottom=2pt, fonttitle=\normalsize, breakable]
\textbf{\textcolor{magenta}{Key Insights for RQ2:}} 
Most existing HNN algorithms, when applied to large-scale datasets, either suffer from efficiency and scalability issues or fail to deliver satisfactory utility. Investigating decoupled architectures that separate high-order information propagation from training modules presents a promising avenue for achieving efficient, scalable, and high-performing HNNs.
\end{tcolorbox}

\subsection{Robustness Evaluation (\textbf{\textcolor{magenta}{RQ3}})}
\label{exp:robustness}

In this section, we assess HNN robustness by simulating structural, feature, and supervision perturbations, as detailed in Appendix~\ref{robust_setups}. 
While our experiments primarily focus on the node classification task due to space limits, DHG-Bench supports flexible extension to other tasks.
We evaluate 10 representative models on four datasets (Cora, Pubmed, Actor, and Pokec). The results on Pubmed and Pokec (Figures~\ref{fig:struc_robust_app},~\ref{fig:feat_robust_app}, and~\ref{fig:super_robust_app}) are provided in Appendix~\ref{exp:add_robust}.

\subsubsection{Robustness Analysis with respect to Structure Perturbations}

%\noindent \textbf{Settings}. To analyze structure-level robustness, following~\citep{cai2022hypergraph}, we randomly remove or add a proportion of node–hyperedge connections in the original hypergraph and then train and evaluate HNN algorithms on the perturbed structures. The modification ratio ranges from 0 to 0.9 to simulate varying levels of noise intensity.

\begin{figure*}[t]
\centering
\includegraphics[width=0.95\linewidth]{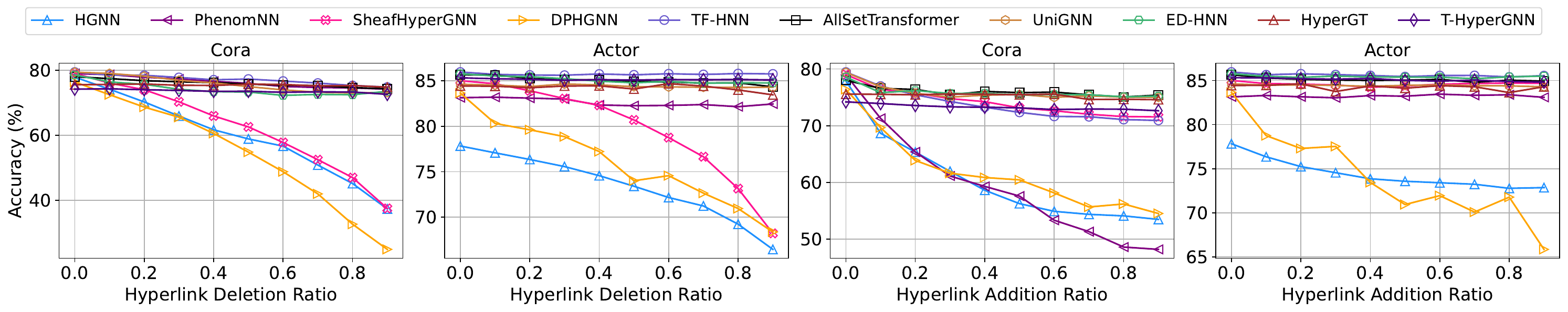}
\caption{Structure robustness analysis on Cora and Actor.}
\label{fig:struc_robust}
\end{figure*}

\noindent \textbf{Results} (Figure~\ref{fig:struc_robust} and Figure~\ref{fig:struc_robust_app}).
\circled{1} Most HNN algorithms exhibit strong robustness against random structural noise, experiencing only marginal performance drops or even remaining nearly unaffected under high perturbation rates. For example, when 90\% of hyperlinks are randomly removed from Cora, 7 out of 10 methods degrade by less than 7\%. Similarly, when 90\% of random hyperlinks are injected into Actor, only 2 models show a noticeable decline in performance.
\circled{2} Spectral-based approaches are generally more vulnerable to structural perturbations. On Pubmed, for instance, increasing the ratio of noisy hyperlinks results in a pronounced performance decline across four spectral-based methods (HGNN, PhenomNN, DPHGNN, and TF-HNN), whereas most other methods remain stable. This may be because spectral methods rely on the hypergraph's global eigenstructure, which is highly sensitive to topological noise.
\circled{3} The robustness of HNN algorithms varies with both the type of structural perturbation (deletion vs. addition) and the choice of dataset. For example, on the Actor dataset, SheafHyperGNN suffers substantial performance degradation under hyperlink deletion but demonstrates strong robustness under hyperlink addition. In another case, PhenomNN exhibits strong robustness on Cora in the addition scenario while showing the opposite trend on Actor.

\subsubsection{Robustness Analysis with respect to Feature Perturbations}

\begin{figure*}[t]
\centering
\includegraphics[width=0.95\linewidth]{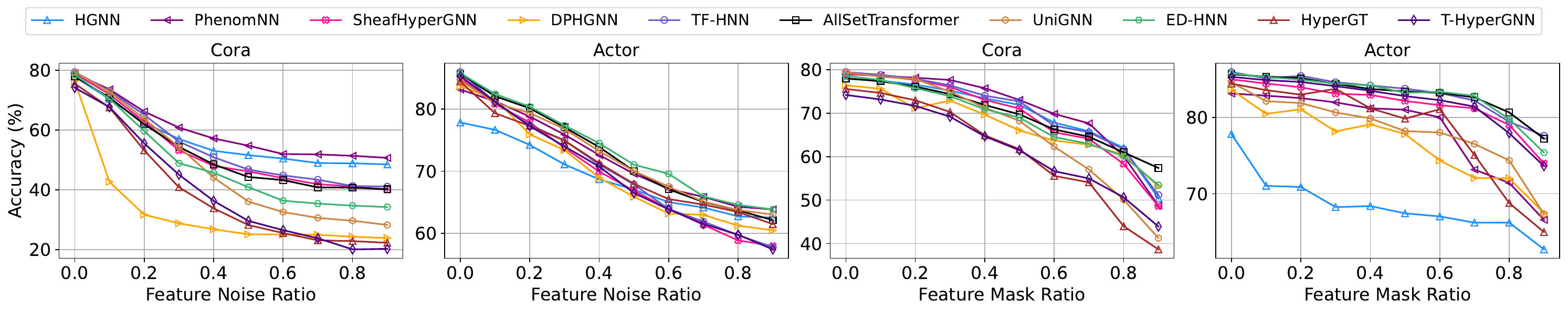}
\caption{Feature robustness analysis on Cora and Actor.}
\label{fig:feat_robust}
\end{figure*}

\begin{figure*}[t!]
\centering
\includegraphics[width=0.95\linewidth]{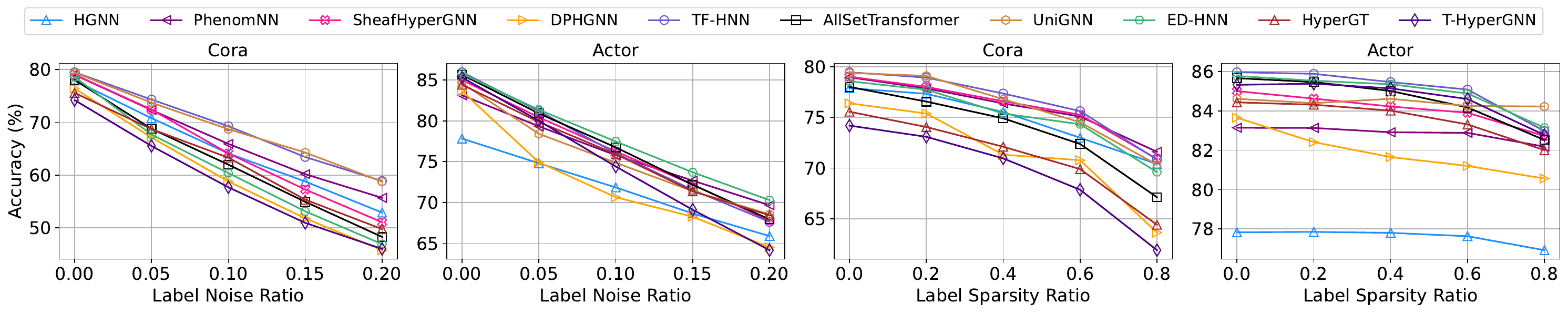}
\caption{Supervision robustness analysis on Cora and Actor.}
\label{fig:super_robust}
\end{figure*}

% \noindent \textbf{Settings}. To study feature-level robustness, we simulate two realistic types of feature perturbations: feature noise and feature sparsity. For feature noise, following~\citep{wu2020graph}, we add independent Gaussian noise to each feature dimension of all nodes with gradually increasing amplitude. Specifically, we use the mean of the maximum feature value of each node as the reference amplitude $r$, and add Gaussian noise $\lambda \cdot r \cdot \epsilon$ to each feature dimension, where $\epsilon \sim \mathcal{N}(0,1)$ and $\lambda$ denotes the feature noise ratio. We evaluate model performance with $\lambda$ varying from 0 to 0.9 at an interval of 0.1. For feature sparsity, following~\citep{li2023gslb}, we randomly mask node features by setting a certain fraction to zero, with the sparsity ratio ranging from 0 to 0.9 at an interval of 0.1.

\noindent \textbf{Results} (Figure~\ref{fig:feat_robust} and Figure~\ref{fig:feat_robust_app}). 
\circled{1} Feature perturbations under equal noise or sparsity levels result in greater performance degradation than structural ones, indicating a more critical role of node features in model prediction.
\circled{2} With increasing noise intensity, model accuracy decreases sharply at the beginning and then stabilizes, as highly corrupted features approximate randomness and lose predictive utility.
\circled{3} As the feature masking rate increases, model performance degrades progressively faster, with a slow decline at low ratios and a sharp drop under high sparsity.
\circled{4} Compared to feature sparsity, feature noise poses a greater challenge for HNN algorithms, with equivalent levels of noise typically resulting in lower predictive accuracy across different datasets.

\subsubsection{Robustness Analysis with respect to Supervision Perturbations}

% \noindent \textbf{Settings}.  We study supervision signal robustness by simulating realistic noise and sparsity scenarios. For label noise, a certain proportion of training samples are randomly assigned incorrect labels, with the noise ratio varying from 0 to 0.20 at intervals of 0.05. Sparsity is introduced by reducing the proportion of training nodes, with the sparsity rate ranging from 0 to 0.8 in increments of 0.2.

\noindent \textbf{Results} (Figure~\ref{fig:super_robust} and Figure~\ref{fig:super_robust_app}). 
\circled{1} As noise intensity increases or supervision becomes sparser, all models show a clear downward trend in performance, with label noise exerting a more pronounced impact.
\circled{2} Increasing label noise generally causes a rapid yet steady decline in performance, which appears approximately linear in most cases.
\circled{3} The impact of supervision sparsity is modest at lower levels but intensifies at higher ratios, resulting in an accelerating decline in model performance. This trend highlights the challenges faced by current HNNs in low-label scenarios.
\circled{4} Label noise and sparsity tend to degrade performance more substantially on homophilic datasets than on heterophilic ones, reflecting the reliance of model predictions on data homophily.

\begin{tcolorbox}[colback=white, colframe=black, sharp corners, boxrule=0.5pt,
  left=2pt, right=2pt, top=2pt, bottom=2pt, fonttitle=\normalsize, breakable]
\textbf{\textcolor{magenta}{Key Insights for RQ3:}} 
Most HNN algorithms demonstrated remarkable robustness to random structural noise, but are considerably more vulnerable to feature perturbations. 
In addition, at the label level, even simple small-scale poisoning attacks can substantially degrade predictive performance, and HNNs face significant challenges under extreme label sparsity.
These findings underscore the need for designing robust HNN architectures or training techniques capable of providing strong defenses against diverse forms of noisy data.
\end{tcolorbox}

\subsection{Fairness Evaluation (\textbf{\textcolor{magenta}{RQ4}})}
\label{exp:fairness}

\begin{figure*}[t]
\centering
\includegraphics[width=0.95\linewidth]{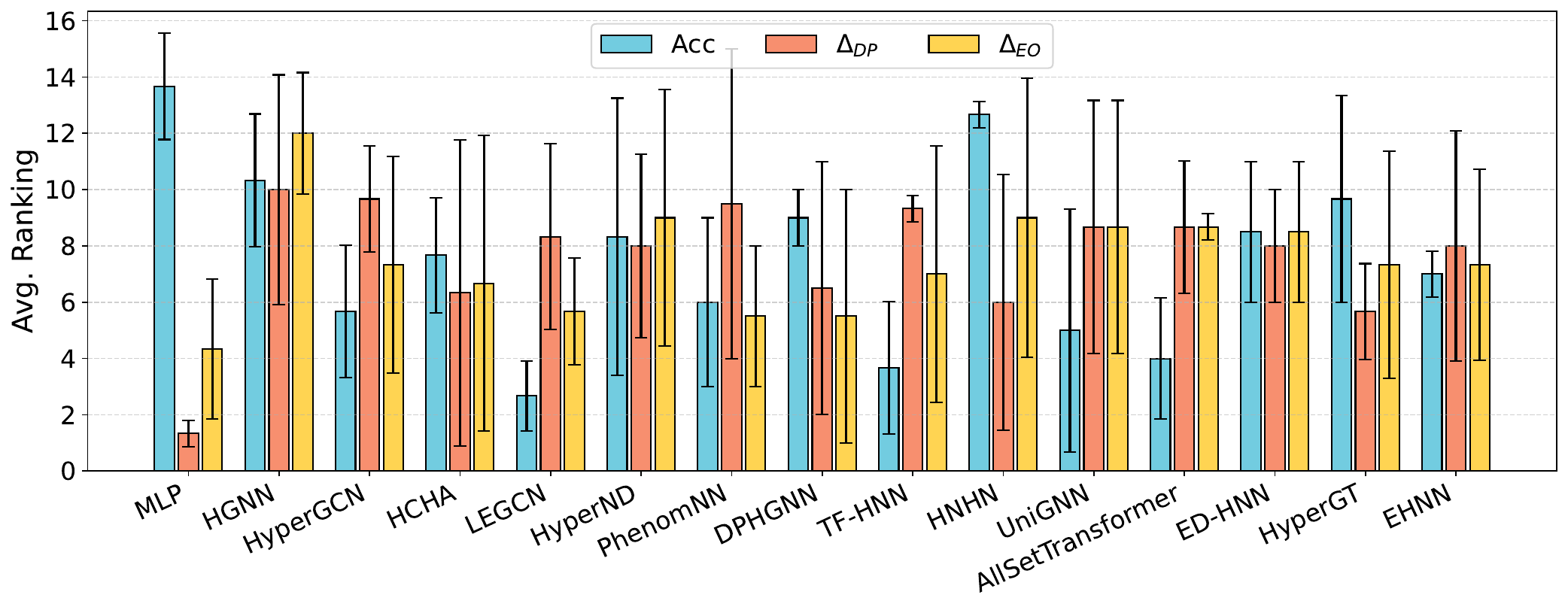}
\caption{Average rankings on Acc, $\Delta_{DP}$, $\Delta_{EO}$ across
the German, Bail, and Credit datasets, where lower values indicate better ranks (ascending order).}
\label{fig:fair_rank}
\end{figure*}

% Methods are ranked in ascending order by the summation of average rankings on all three metrics.

In this section, we analyze algorithmic fairness and report full quantitative results in terms of accuracy (Acc), $\Delta_{DP}$, and $\Delta_{EO}$ in Table~\ref{tab:fairness} of Appendix~\ref{exp:add_fairness}. To better illustrate the strengths and limitations of each algorithm, we present Figure~\ref{fig:fair_rank}, which shows their average rankings across the three metrics on datasets where they can run, considering only HNNs executable on at least two datasets.

\begin{comment}
In this section, we analyse algorithmic fairness and report full quantitative results in terms of model utility (Acc) and fairness ($\Delta_{DP}$ and $\Delta_{EO}$) in Table~\ref{tab:fairness} in Appendix~\ref{exp:add_fairness}.
To better understand the strengths and limitations of each algorithm, we calculate their average rankings across three metrics on datasets where they can run, considering only HNNs executable on at least two datasets, as shown in Figure~\ref{fig:fair_rank}.
\end{comment}

\noindent \textbf{Results} (Figure~\ref{fig:fair_rank} and Table~\ref{tab:fairness}).
\circled{1} While HNN algorithms achieve higher predictive performance, they generally suffer from more severe fairness issues compared to MLP, which is free from message passing. 
Figure~\ref{fig:fair_rank} shows that MLP ranks best on the two fairness metrics but worst on accuracy. 
For example, on the Credit dataset, MLP achieves lower $\Delta_{DP}$ and $\Delta_{EO}$ values than HCHA, the fairest among the evaluated HNN models, as shown in Table~\ref{tab:fairness}.
%For example, Table~\ref{tab:fairness} shows that on the Credit dataset, MLP achieves lower values on both $\Delta_{DP}$ and $\Delta_{EO}$ compared to HCHA, the fairest among the evaluated HNN models.
%MLP attains a $\Delta_{DP}$ of 3.43 and a $\Delta_{EO}$ of 2.07 on the Credit dataset, both lower than those of HCHA, the fairest HNN model.
%On the Credit dataset, MLP achieves $\Delta_{DP}$ of 3.43 and $\Delta_{EO}$ of 2.07, both lower than those of HCHA, the fairest HNN model.
\circled{2} The fairness performance of HNN algorithms varies considerably across datasets, with no method achieving consistently superior performance on all benchmarks. For instance, Table~\ref{tab:fairness} illustrates that while HCHA achieves the best fairness performance on the Credit dataset across both metrics, its $\Delta_{DP}$ and $\Delta_{EO}$ rank as the second- and third-worst, respectively, on the German dataset. Moreover, Figure~\ref{fig:fair_rank} shows that most algorithms exhibit substantial variance in their rankings, further highlighting the instability of fairness across datasets.
\circled{3} HNN algorithms show inconsistent behavior across fairness metrics, and strong performance on one does not guarantee superiority on another. For example, on the Bail dataset, although HNHN achieves the lowest $\Delta_{DP}$ among all HNN methods, its $\Delta_{EO}$ ranks as the third worst among the 17 HNN models.

\begin{tcolorbox}[colback=white, colframe=black, sharp corners, boxrule=0.5pt,
  left=2pt, right=2pt, top=2pt, bottom=2pt, fonttitle=\normalsize, breakable]
\textbf{\textcolor{magenta}{Key Insights for RQ4:}} 
Existing HNN algorithms tend to produce more biased predictions than MLPs, indicating that high-order information propagation may exacerbate the amplification of biases from sensitive information. Moreover, fairness performance varies substantially across datasets and metrics.
These findings highlight the need for developing debiased algorithms that can achieve stronger fairness across diverse high-stakes real-world applications.
\end{tcolorbox}

\section{Conclusion}

This paper introduces DHG-Bench, the first comprehensive benchmark for deep hypergraph learning, which integrates and compares 17 representative HNNs across 22 hypergraph datasets encompassing various domains, sizes, and structural properties, under consistent experimental settings. We comprehensively evaluate the effectiveness, efficiency, robustness, and fairness of HNN algorithms, and our analysis reveals the strengths and weaknesses of different HNNs in a wide range of scenarios, offering valuable insights into their practical applicability and design trade-offs.
Furthermore, we develop and release a package, DHG-Bench, that includes all experimental protocols, baseline algorithms, datasets, and reproducibility scripts to facilitate future research. 
To foster further advances in deep hypergraph learning, we outline several promising future directions for HNNs in Appendix~\ref{future_direct}.

\bibliography{iclr2026_conference}
\bibliographystyle{iclr2026_conference}

\clearpage 
\appendix

\renewcommand{\thefigure}{A\arabic{figure}}
\renewcommand{\theHfigure}{A\arabic{figure}} % <- 关键
\setcounter{figure}{0}

\renewcommand\thetable{A\arabic{table}}    
\setcounter{table}{0} 

\section*{\textbf{Appendix}}

\section{Datasets and Algorithms}

\subsection{Benchmark Datasets}
\label{datasets_desc}

\begin{table*}[h!]
    \centering
    \caption{Statistics of the standard node-level datasets: $|e|$ denotes the hyperedge size, while $\mathcal{H}_{\text{edge}}$ indicates the hyperedge homophily ratio introduced in~\citep{li2025hypergraph}.}
    \resizebox{0.9\textwidth}{!}{%
    \begin{tabular}{l c c c c c c}
        \toprule
        \textbf{Dataset} & \# \textbf{Nodes} & \# \textbf{Edges} & \# \textbf{Features} & \textbf{Avg.} $|e|$ & $\mathcal{H}_{\text{edge}}$ & \# \textbf{Classes} \\
        \midrule
        Cora        & 2,708   & 1,579   & 1,433  & 3.03  & 0.75 & 7 \\
        Pubmed      & 19,717  & 7,963   & 500    & 4.35  & 0.78 & 3 \\
        Cora-CA     & 2,708   & 1,072   & 1,433  & 4.28  & 0.78 & 7 \\
        DBLP-CA     & 41,302  & 22,363  & 1,425  & 4.45  & 0.87 & 6 \\
        NTU2012     & 2,012   & 2,012   & 100    & 5.00  & 0.79 & 67 \\
        ModelNet40  & 12,311  & 12,311  & 100    & 5.00  & 0.87 & 40 \\
        Walmart     & 88,860  & 69,906  & 100    & 6.59  & 0.60 & 11 \\
        Trivago     & 172,738 & 233,202 & 300    & 3.12  & 0.98 & 160 \\
        \midrule
        Actor       & 16,255  & 10,164  & 50     & 5.25  & 0.46 & 3 \\
        Ratings     & 22,299  & 2,090   & 111    & 3.10  & 0.37 & 5 \\
        Gamers      & 16,812  & 2,627   & 7      & 6.23  & 0.49 & 2 \\
        Pokec       & 14,998  & 2,406   & 65     & 2.29  & 0.45 & 2 \\
        Yelp        & 50,758  & 679,302 & 1,862  & 6.66  & 0.29 & 9 \\
        \bottomrule
    \end{tabular}}
    \label{tab:dataset-full}
\end{table*}

\begin{comment}
\begin{table*}[h!]
    \centering
    \caption{Statistics of the standard node-level datasets: $|e|$ denotes the hyperedge size, while $\mathcal{H}_{\text{edge}}$ indicates the hyperedge homophily ratio.}
    \resizebox{0.9\textwidth}{!}{%
    \begin{tabular}{l c c c c c c}
        \toprule
        \textbf{Dataset} & \# \textbf{Nodes} & \# \textbf{Edges} & \# \textbf{Features} & \# \textbf{Classes} & \textbf{Avg.} $|e|$ & $\mathcal{H}_{edge}$ \\
        \midrule
        Cora        & 2,708   & 1,579   & 1,433  & 7   & 3.03  & 1.77 \\
        Pubmed      & 19,717  & 7,963   & 500   & 3   & 4.34  & 1.86 \\
        Cora-CA     & 2,708   & 1,072   & 1,433  & 7   & 4.28  & 1.69 \\
        DBLP-CA     & 41,302  & 22,363  & 1,425  & 6   & 4.45  & 2.41 \\
        NTU2012     & 2,012   & 2,012   & 100   & 67  & 5.00  & 5.00 \\
        ModelNet40  & 12,311  & 12,311  & 100   & 40  & 5.00  & 5.00 \\
        Walmart     & 88,860  & 69,906  & 100   & 11  & 6.59  & 5.18 \\
        Trivago     & 172,738 & 233,202 & 300   & 160 & 3.12  & 4.21 \\
        \midrule
        Actor       & 16,255  & 10,164  & 50    & 3   & 5.25  & 3.28 \\
        Ratings     & 22,299  & 2,090   & 111   & 5   & 4.00  & 2.92 \\
        Gamers      & 18,612  & 2,627   & 7     & 2   & 6.23  & 2.97 \\
        Pokec       & 14,998  & 2,406   & 65    & 2   & 2.29  & 2.77 \\
        Yelp        & 50,58  & 679,302 & 1,862  & 9   & 6.66  & 89.12 \\
        \bottomrule
    \end{tabular}}
    \label{tab:dataset-full}
\end{table*}
\end{comment}

\begin{table*}[h!]
    \centering
    \caption{Statistics of fairness-sensitive datasets. \textbf{Sens} denotes the sensitive attribute.}
    \resizebox{0.9\textwidth}{!}{\begin{tabular}{l c c c c c}
        \toprule
    \textbf{Dataset} & \# \textbf{Nodes} & \# \textbf{Edges} & \# \textbf{Features} & \textbf{Sens} & \textbf{Label} \\
    \midrule
    German 	& 1,000	&  1,000 & 27	& Gender &  Credit status \\
    Bail	& 18,876	& 18,876	&	18 & Race	& Bail decision \\
    Credit	& 30,000	& 30,000	& 13	& Age	& Future default\\
        \bottomrule
    \end{tabular}}
    \label{tab:fairstat}
\end{table*}

\begin{table*}[h!]
    \centering
    \caption{Statistics of graph-level datasets. Avg. $|\mathcal{V}|$, $|\mathcal{E}|$, and $|e|$ represent the average number of nodes, hyperedges, and hyperedge sizes, respectively.}
    \resizebox{0.9\textwidth}{!}{\begin{tabular}{l c c c c c}
        \toprule
        \textbf{Dataset} & \# \textbf{Hypergraphs} & \textbf{Avg. $|\mathcal{V}|$} & \textbf{Avg. $|\mathcal{E}|$} & \textbf{Avg. $|e|$} & \# \textbf{Classes}\\
        \midrule
       RHG-10	& 2,000	& 31.3	& 29.8	& 5.2 & 10 \\
       RHG-3	& 1,500	& 35.5	&	17.9 & 6.9	& 3 \\
       IMDB-Dir-Form	& 1,869	& 15.7	& 39.2	& 3.7	& 3\\
       IMDB-Dir-Genre & 3,393	& 17.3	&	36.4 & 3.8	& 3 \\
        Steam-Player	& 2,048	& 13.8	&	46.4 & 4.5	& 2 \\
        Twitter-Friend & 1,310	& 21.6	& 84.3	& 4.3 & 2 \\
        \bottomrule
    \end{tabular}}
    \label{tab:dataset_hypergraph}
\end{table*}

We adopt 22 publicly available benchmark datasets to comprehensively evaluate HNN algorithms. The statistics of node-level datasets, fairness-sensitive datasets, and graph-level datasets are reported in Tables~\ref{tab:dataset-full},~\ref{tab:fairstat}, and~\ref{tab:dataset_hypergraph}, respectively.
Detailed descriptions of these datasets are provided below.

\begin{itemize}[leftmargin=15pt]
    \item \textbf{Cora/Pubmed/Cora-CA/DBLP-CA}~\citep{yadati2019hypergcn}: Cora and Pubmed are co-citation networks where nodes represent papers and hyperedges connect papers cited together. Cora-CA and DBLP-CA are co-authorship hypergraphs, with nodes as papers and hyperedges linking all papers co-authored by the same author. Node features are Bag-of-Words (BoW)~\citep{zhang2010understanding} representations of the documents, and labels indicate paper categories.
    \item \textbf{NTU2012/ModelNet40}~\citep{feng2019hypergraph}: The ModelNet40 and the NTU2012 are two computer vision and graphics datasets. ModelNet40 contains 12,311 3D objects from 40 popular categories, while NTU2012 consists of 2,012 3D shapes from 67 categories. For each object, features are extracted using both the Group-View Convolutional Neural Network (GVCNN)\citep{fenggroup2018} and the Multi-View Convolutional Neural Network (MVCNN)\citep{su2015multi}. Following~\citep{feng2019hypergraph}, we construct hyperedges by aggregating the nearest neighbors of each node based on Euclidean distance.
    %\item \textbf{Zoo}~\citep{Dua:2017} The Zoo dataset is modeled as a hypergraph, where each node corresponds to an animal, and hyperedges are formed by grouping animals that share the same attributes (e.g., hair, feathers, number of legs). Node labels indicate the animal categories, including mammals, birds, reptiles, fish, amphibians, insects, and invertebrates. Each node feature is expressed as a 17-dimensional binary vector that encodes the presence or absence of these attributes.
    %\item \textbf{House/Walmart}~\citep{chien2022you} The House dataset is represented as a hypergraph, where each node corresponds to a member of the U.S. House of Representatives, and hyperedges are constructed by grouping members who serve on the same committee. Node labels denote the political party affiliation of the representatives. The Walmart dataset models a hypergraph where nodes represent products and hyperedges capture sets of products purchased together. Node labels indicate product categories. Following~\citep{chien2022you}, in both datasets, each node feature is a 100-dimensional vector obtained by adding Gaussian noise $\mathcal{N}(0,\sigma^{2}\mathrm{I})$ with $\sigma=0.6$ to one-hot encodings of the labels.
    \item \textbf{Walmart}~\citep{chien2022you}: The Walmart dataset models a hypergraph where nodes represent products and hyperedges capture sets of products purchased together. Node labels indicate product categories. Following~\citep{chien2022you}, each node feature is a 100-dimensional vector obtained by adding Gaussian noise $\mathcal{N}(0,\sigma^{2}\mathrm{I})$ with $\sigma=0.6$ to one-hot encodings of the labels.
    \item \textbf{Trivago}~\citep{kim2023datasets}: Trivago is a hotel-web search hypergraph where each node indicates a hotel, and each hyperedge corresponds to a user. If a user (hyperedge) has 
visited the website of a particular hotel (node), the corresponding node is added to 
the respective user hyperedge. Furthermore, each hotel's class is labeled based on 
the country in which it is located. 
    \item \textbf{Actor}~\citep{li2025hypergraph}: The actor co-occurrence network is derived from a heterogeneous movie-actor-director-writer network~\footnote{\url{https://www.aminer.org/lab-datasets/soinf/}}, capturing intricate collaborations within films. Nodes represent individuals involved in film production (actors, directors, and writers), and hyperedges denote their joint participation in a single film. Node attributes are extracted from Wikipedia keywords, and labels indicate each individual's specific role.
    \item \textbf{Amazon-ratings (Ratings)}~\citep{li2025hypergraph}: This dataset, sourced from the Amazon co-purchasing network in the SNAP repository~\citep{snap}, includes products like books, music CDs, DVDs, and VHS tapes. Nodes represent individual products, and hyperedges link those frequently purchased together. The task is to predict each product's average user rating, classified into ten levels. Node features are extracted using the BoW technique applied to product descriptions~\citep{juluru2021bag}.
    \item \textbf{Twitch-gamers (Gamers)}~\citep{li2025hypergraph}: The Twitch-gamers dataset is a connected undirected hypergraph representing user interactions on the Twitch streaming platform. Nodes denote user accounts, and hyperedges are formed based on mutual follows within specific timeframes. Each node is associated with features such as view counts, timestamps, language preferences, activity duration, and inactivity status. The goal is to predict whether a channel hosts explicit content (binary classification).
    \item \textbf{Pokec}~\citep{li2025hypergraph}: The Pokec dataset is derived from Slovakia's largest online social networking platform and is used to model social relationships and attributes. Nodes represent individual users, and hyperedges correspond to each user’s full set of friends. Node labels indicate user-reported gender, while node features are extracted from profile information, including age, hobbies, interests, education level, region, etc.
    \item \textbf{Yelp}~\citep{chien2022you}: The Yelp dataset is a hypergraph where nodes represent restaurants and hyperedges link those visited by the same user. Node labels denote average star ratings (1.0–5.0 in 0.5 steps). Features include geographic coordinates, one-hot encodings of city/state, and BoW vectors from the top-1000 restaurant name tokens.
    \item \textbf{German}~\citep{wei2022augmentations}: The nodes in the dataset represent clients in the German Bank, and hyperedges are constructed by
linking individuals with the most similar credit accounts
to each person in the dataset. The task is to classify credit
risk levels as high or low based on the sensitive attribute
”gender” (Male/Female).
    \item \textbf{Bail}~\citep{wei2022augmentations}: The nodes in the datasets are defendants who got
released on bail at the U.S state courts during 1990-
2009~\citep{jordan2015effect}. Hyperedges are constructed based on the
similarity of past criminal records among individuals. The
task is to classify whether defendants are on bail or not
with the sensitive attribute ”race” (White/Black).
    \item \textbf{Credit}~\citep{wei2022augmentations}: The nodes in the dataset represent credit
card users, and hyperedges are formed based on the similarity of users' spending and payment patterns. The
task is to classify the default status with the sensitive
attribute ”age” ($<$25 / $>$25). 
  \item \textbf{RHG-10/RHG-3}~\citep{feng2024hypergraph}: RHG-10 dataset encompasses ten distinct synthetic factor hypergraph structures
  (i.e., Hyper Flower, Hyper Pyramid, Hyper Checked Table, Hyper Wheel, Hyper Lattice, Hyper Windmill, Hyper Firm Pyramid, Hyper RChecked Table,
  Hyper Cycle, and Hyper Fern). To evaluate the algorithm's ability to recognize significant high-order structures, the RHG-3 dataset is constructed by randomly generating
 hypergraphs for three distinctively various hypergraph structures: Hyper Pyramid, Hyper Checked Table, and Hyper Wheel.

  \item \textbf{IMDB-Dir-Form/IMDB-Dir-Genre}~\citep{feng2024hypergraph}: These two datasets contain hypergraphs constructed by the 
  co-director relationship from the original IMDB dataset. The director of each movie is a hypergraph. "Form" included in the dataset's name indicates that the
  movie category is identified by its form, like animation, documentary, and drama. "Genre" denotes that the movie is classified by its genres, like adventure, crime, and family. 
  \item \textbf{Steam-Player}~\citep{feng2024hypergraph}: The Steam-Player dataset is a player dataset where each player is a hypergraph. The vertex is
  the games played by the player, and the hyperedge is constructed
  by linking the games with shared tags. The target of the dataset
  is to identify each user's preference: single-player game or
  multiplayer game.
  \item \textbf{Twitter-Friend}~\citep{feng2024hypergraph}: The Twitter-Friend dataset is a social media
  dataset. Each hypergraph is the friends of a specified user. The
  hyperedge is constructed by linking the users who are friends.
  The label associated with the hypergraph is to identify whether
  the user posted the blog about "National Dog Day" or "Respect
  Tyler Joseph".
\end{itemize}

\begin{figure*}[t]
\centering
\includegraphics[width=1\linewidth]{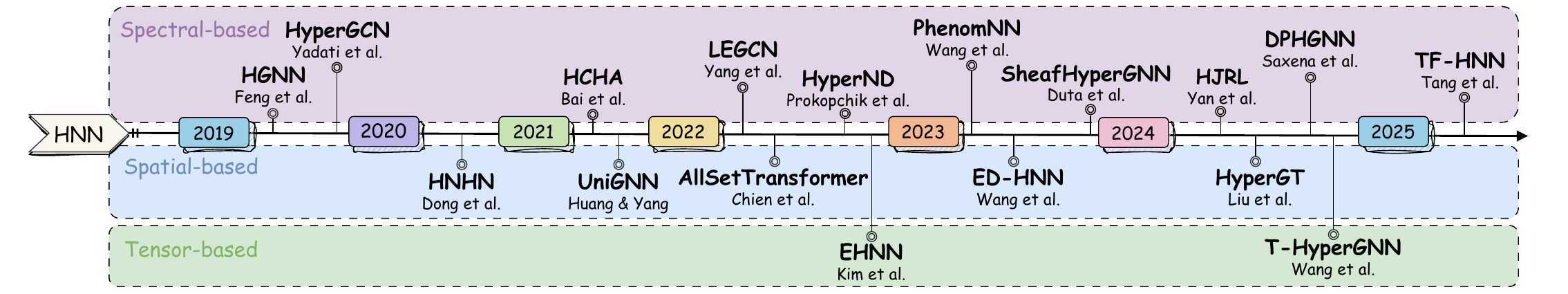}
\caption{A timeline of the representative hypergraph neural networks.}
\label{fig:timeline}
\end{figure*}

\subsection{Benchmark Algorithms}
\label{algo_details}

Figure~\ref{fig:timeline} illustrates 17 HNN algorithms integrated into our DHG-Bench, including 10 spectral-based, 5 spatial-based, and 2 tensor-based methods.
We introduce these methods in detail below.

\subsubsection{Spectral-Based Algorithms}

\begin{itemize}[leftmargin=15pt]
  \item \textbf{HGNN}~\citep{feng2019hypergraph}: HGNN is a framework for representation learning that extends spectral convolution to hypergraphs. By leveraging the hypergraph Laplacian and approximating spectral filters with truncated Chebyshev polynomials, it effectively captures high-order correlations inherent in complex data. 
  \item \textbf{HyperGCN}~\citep{yadati2019hypergcn}: HyperGCN approximates each hyperedge of the hypergraph by a set of pairwise edges connecting the vertices of the hyperedge, and treats the learning problem as a graph learning task on the approximated graph.
  \item \textbf{HCHA}~\citep{bai2021hypergraph}: HCHA is a hypergraph neural network that introduces two end-to-end trainable operators: hypergraph convolution and hypergraph attention. Hypergraph convolution efficiently propagates information by leveraging high-order relationships and local clustering structures, with standard graph convolution shown as a special case. Hypergraph attention further enhances representation learning by dynamically adjusting hyperedge connections through an attention mechanism, enabling task-relevant information aggregation and yielding more discriminative node embeddings.
  \item \textbf{LEGCN}~\citep{yang2022semi}: LEGCN is a hypergraph learning model based on the Line Expansion (LE). By modeling vertex-hyperedge pairs, LEGCN bijectively transforms a hypergraph into a simple graph, preserving the symmetric co-occurrence structure and avoiding information loss. This enables existing graph learning algorithms to operate directly on hypergraphs. 
  \item \textbf{HyperND}~\citep{prokopchik2022nonlinear}: HyperND develops a
 nonlinear diffusion process on hypergraphs that
 propagates both features and labels along the
 hypergraph structure. The novel diffusion
 incorporates a broad class of nonlinearities to increase
 the modeling capability, and the limiting point serves
 as a node embedding from which we make predictions with a linear model.
  \item \textbf{PhenomNN}~\citep{wang2023hypergraph}: PhenomNN is a hypergraph learning framework grounded in a family of expressive, parameterized hypergraph-regularized energy functions. It formulates node embeddings as the minimizers of these energy functions, which are optimized jointly with a parameterized classifier through a supervised bilevel optimization process. This approach provides a principled way to model high-order relationships in hypergraphs while enabling end-to-end training. 
  \item \textbf{SheafHyperGNN}~\citep{duta2023sheaf}: SheafHyperGNN introduces a cellular sheaf framework for hypergraphs, enabling the modeling of complex dynamics while preserving their higher-order connectivity. Then, it generalizes the two
 commonly used hypergraph Laplacians to incorporate the richer structure sheaves offer and constructs two powerful neural networks capable of inferring and processing hypergraph sheaf structure.
  \item \textbf{HJRL}~\citep{yan2024hypergraph}: HJRL introduces a novel cross
 expansion method, which transforms both hypervertices and edges of a hypergraph to vertices in a standard graph. Then, a joint learning model is proposed to embed both hypervertices and hyperedges into a shared representation space.
 In addition, the algorithm employs a hypergraph reconstruction objective to preserve structural information in the model.
 \item \textbf{DPHGNN}~\citep{saxena2024dphgnn}: DPHGNN is a hybrid framework designed for effective feature representation in resource-constrained hypergraph settings. It introduces equivariant operator learning to capture lower-order semantics by inducing topology-aware inductive biases. It employs a dual-layered feature update mechanism: a static update layer provides spectral biases and relational features, while a dynamic update layer fuses explicitly aggregated features from the underlying topology into the hypergraph message-passing process.
 \item \textbf{TF-HNN}~\citep{tangtraining}: TF-HNN is the first model to decouple hypergraph structural processing from model training, substantially improving training efficiency. Specifically, it introduces a unified, training-free message-passing module (TF-MP-Module) by identifying feature aggregation as the core operation in HNNs. The TF-MP-Module removes learnable parameters and nonlinear activations, and compresses multi-layer propagation into a single step, offering a simplified and efficient alternative to existing architectures.
\end{itemize}

\subsubsection{Spatial-Based Algorithms}

\begin{itemize}[leftmargin=15pt]
  \item \textbf{HNHN}~\citep{dong2020hnhn}: HNHN is a hypergraph convolution network with nonlinear activation
 functions applied to both hypernodes and hyperedges, combined with a normalization scheme
 that can flexibly adjust the importance of high-cardinality hyperedges and high-degree vertices
 depending on the dataset. 
  \item \textbf{UniGNN}~\citep{huang2021unignn}: UniGNN is a unified message-passing framework that generalizes standard GNNs to hypergraphs. It models the two-stage aggregation process by first computing hyperedge representations using a permutation-invariant function over the features of incident vertices, and then updating each vertex by aggregating its associated hyperedge representations. This formulation enables seamless adaptation of existing GNN architectures to hypergraph structures.
  \item \textbf{AllSetTransformer}~\citep{chien2022you}: AllSetTransformer is a novel HNN paradigm that implements each layer as a composition of two multiset functions. By incorporating the Set Transformer~\citep{lee2019set} into its architecture, it achieves greater modeling flexibility and enhanced expressive power.
  %AllSetTransformer is a novel HNN paradigm that implements HNN layers as compositions of two multiset functions. The model integrates a powerful Set Transformer~\citep{lee2019set} in the design of the specialized HNN, enabling greater modeling flexibility and enhanced expressive power.
  \item \textbf{ED-HNN}~\citep{wang2023equivariant}: ED-HNN is an architecture designed to approximate any continuous, permutation-equivariant hypergraph diffusion operator. The model is efficiently implemented by combining the star expansion (bipartite representation) of hypergraphs with standard message-passing neural networks, and supports scalable training via shared weights across layers.
  \item \textbf{HyperGT}~\citep{liu2024hypergraph}: HyperGT is a Transformer-based HNN architecture designed to capture global correlations among nodes and hyperedges. To preserve local structural information, it incorporates incidence-matrix-based positional encoding and a structure regularization term. These designs enable comprehensive hypergraph representation learning by jointly modeling global interactions and local connectivity patterns.
\end{itemize}

\subsubsection{Tensor-Based Algorithms}

\begin{itemize}[leftmargin=15pt]
  \item \textbf{EHNN}~\citep{kim2022equivariant}: EHNN is the first framework to realize equivariant GNNs for general hypergraph learning. It establishes a connection between sparse hypergraphs and dense, fixed-order tensors, enabling the design of a maximally expressive equivariant linear layer. To ensure scalability and generalization to arbitrary hyperedge orders, EHNN further introduces hypernetwork-based parameter sharing.
  \item \textbf{T-HyperGNN}~\citep{wang2024t}: T-HyperGNN is a general framework that integrates tensor hypergraph signal processing (t-HGSP)~\citep{pena2023t} to encode hypergraph structures using tensors. It models node interactions through multiplicative interaction tensors, elevating aggregation from traditional linear operations to higher-order polynomial mappings, thereby enhancing expressive power. To ensure scalability, T-HyperGNN introduces tensor-message-passing by exploiting tensor sparsity, enabling efficient processing of large hypergraphs with computational and memory costs comparable to matrix-based HNNs.
\end{itemize}

In addition, we include MLP and two GNN-based methods, CEGCN and CEGAT~\citep{chien2022you}, as baselines in our comparative study. Both CEGCN and CEGAT are expansion-based approaches that transform a hypergraph into a pairwise graph via clique expansion~\citep{zhou2006learning}, where each hyperedge is converted into a clique over its incident nodes. Specifically, CEGCN applies GCN~\citep{kipf2017semisupervised} to the expanded graph, while CEGAT employs GAT~\citep{velivckovic2018graph} to model node importance within the cliques.
\section{Details of the Experimental Settings}
\label{exp_details}

\subsection{General Experimental Settings}
\label{exp_setups}

\begin{table*}[tbp]
\centering
\caption{Hyperparameter search space of different methods.}
\renewcommand{\arraystretch}{1.0}
\begin{tabular}{>{\raggedright}p{3.3cm} >{\raggedright}p{4cm} >{\raggedright\arraybackslash}p{7.2cm}}
\toprule
\textbf{Method} & \textbf{Hyperparameter} & \textbf{Search Space} \\
\midrule
\multirow{9}{*}{\centering General Settings} 
    & Epochs & 100, 200, 300, 400, 500, 800, 1000 \\
    & Learning Rate & 0.1, 0.01, 0.001, 0.0001 \\
    & Layers & 1, 2, 3, 4 \\
    & Dropout Rate & 0, 0.1, 0.2, 0.3, 0.4, 0.5, 0.6, 0.7, 0.8 \\
    & Weight Decay & 0, 0.0005 \\
    & Hidden Units & 64, 128, 256, 512, 1024 \\ 
    & Activation & LeakyReLU, ReLU, PReLU, Sigmoid, Softmax \\
    & Hyperedge Pooling & max, mean, max-min\\
    & Hypergraph Pooling & max, mean \\
    \midrule
\multirow{1}{*}{\centering HCHA}     
    & heads & 1, 2, 4, 8, 16 \\ \midrule
\multirow{4}{*}{\centering HyperND}     
    & HyperND\_ord & 1, 2, 3, 5, 10 \\  
    & HyperND\_tol & 0.001, 0.0001, 0.00001, 0.000001 \\
    & HyperND\_steps & 50, 100, 150, 200 \\
    & alpha & 0, 0.1, 0.2, 0.3, 0.4, 0.5, 0.6, 0.7, 0.8, 0.9 \\
    \midrule 
\multirow{1}{*}{\centering HJRL}     
    & $\lambda_{0}$ & 0.001, 0.01, 0.1, 1, 10 \\ \midrule
\multirow{4}{*}{\centering PhenomNN} 
    & $\lambda_{0}$ & 0, 0.1, 1, 10, 20, 50, 80, 100 \\    
    & $\lambda_{1}$ & 0, 0.1, 1, 10, 20, 50, 80, 100 \\ 
    & prop\_steps & 2, 4, 8, 16, 32, 64, 128 \\
    & alpha & 0, 0.1, 0.2, 0.3, 0.4, 0.5, 0.6, 0.7, 0.8, 0.9 \\
    \midrule 
\multirow{4}{*}{\centering SheafHyperGNN}     
    & init\_hedge & rand, avg \\  
    & sheaf\_pred\_block & MLP\_var1, MLP\_var2, MLP\_var3, cp\_decomp \\
    & sheaf\_transformer\_head & 1, 2, 4, 8, 16 \\
    & stalk\_dim & 1, 2, 4, 8 \\
    \midrule 
\multirow{2}{*}{\centering TF-HNN}     
    & mlp\_hidden\_size & 64, 128, 256, 512, 1024\\  
    & \# layers of classifier & 1, 2, 3, 4\\ \midrule 
\multirow{2}{*}{\centering HNHN}     
    & alpha & -3.0, -2.5, -2.0, -1.5, -1.0, -0.5, 0.0, 0.5 \\  
    & beta & -2.5, -2.0, -1.5, -1.0, -0.5, 0.0, 0.5, 1.0 \\ \midrule 
\multirow{2}{*}{\centering UniGNN}     
    & alpha & 0, 0.1, 0.2, 0.3, 0.4, 0.5, 0.8, 0.9 \\  
    & beta & 0, 0.1, 0.2, 0.3, 0.4, 0.5, 0.8, 0.9 \\ \midrule 
\multirow{1}{*}{\centering AllSetTransformer}     
    & attention\_heads & 1, 2, 4, 8, 16 \\ \midrule
\multirow{4}{*}{\centering ED-HNN}     
    & alpha & 0, 0.1, 0.2, 0.3, 0.4, 0.5, 0.6, 0.7, 0.8, 0.9 \\  
    & \# layers of $\hat{\phi}$ & 0, 1, 2, 3 \\
    & \# layers of $\hat{\rho}$ & 0, 1, 2, 3 \\
    & \# layers of $\hat{\varphi}$ & 0, 1, 2, 3 \\
    \midrule 
\multirow{4}{*}{\centering DPHGNN}     
    & attention\_heads & 1, 2, 4, 8, 16 \\ 
    & \# layers of $\mathrm{TAA}$ module & 1, 2, 3, 4 \\
    & \# layers of $\mathrm{SIB}$ module & 1, 2 \\
    & \# layers of $\mathrm{DFF}$ module & 1, 2  \\
    \midrule
\multirow{1}{*}{\centering HyperGT}     
    & attention\_heads & 1, 2, 4, 8, 16 \\ \midrule 
\multirow{5}{*}{\centering EHNN}     
    & ehnn\_qk\_channels & 64, 128, 256 \\  
    & ehnn\_n\_heads & 1, 2, 4, 8, 16 \\
    & ehnn\_pe\_dim & 64, 128\\
    & ehnn\_inner\_channel & 64, 128, 256 \\
    & ehnn\_hidden\_channel & 64, 128, 256 \\
    \midrule 
\multirow{2}{*}{\centering T-HyperGNN}     
    & M: maximum cardinality & 1, 2, 3, 4, 5 \\  
    & combine & concat, sum \\ 
\bottomrule
\end{tabular}
\label{tab:search space}
\end{table*}

We strive to follow the original implementations of various HNN methods from their respective papers or source codes and integrate them into a unified training and evaluation framework.
All parameters are randomly initialized. We use the cross-entropy loss function~\citep{mao2023cross} for all three benchmark classification tasks. Adam optimizer~\citep{kingma2014adam} is adopted with an appropriate learning rate and weight decay to achieve the best performance on the validation split.
Detailed hyperparameter settings and experimental environments are provided in Appendix~\ref{param_setups} and Appendix~\ref{comp_env}, respectively.
For evaluation, we follow prior studies in choosing task-specific metrics: accuracy for node classification~\citep{feng2019hypergraph,chien2022you,wang2023equivariant}; AUROC (area under the ROC curve) and AP (average precision) for hyperedge prediction~\citep{hwang2022ahp,ko2025enhancing,yu2025hygen,tangtraining}; and both accuracy and Macro-F1 score for hypergraph classification~\citep{feng2024hypergraph}.
Higher values of these metrics indicate better predictive performance. 
In addition, to assess algorithmic fairness, we adopt two commonly used group fairness metrics: demographic parity ($\Delta_{DP}$)~\citep{dwork2012fairness} and equalized odds ($\Delta_{EO}$)~\citep{hardt2016equality}, with detailed definitions provided in Appendix~\ref{eval_metrics}.
For each method and dataset, we record the mean results and the standard deviation across 5 runs.

\subsection{Hyperparameter Setting}
\label{param_setups}

We carefully tune hyperparameters to ensure a rigorous and unbiased evaluation of the integrated HNN methods. For algorithms without explicit hyperparameter guidelines in their original papers or source code, we perform a grid search with a reasonable budget across all datasets to identify optimal configurations. The search spaces are provided in Table~\ref{tab:search space}. For detailed interpretations, please refer to the corresponding papers, and the complete hyperparameter configurations are available in our publicly released GitHub repository.

\subsection{Experimental Environment}
\label{comp_env}

All the experiments are conducted with the following computational resources and configurations:

\begin{itemize}[leftmargin=15pt]
    \item Operating system: Ubuntu 24.04 LTS.
    \item CPU information: Intel(R) Xeon(R) Silver 4208 CPU @ 2.10GHz with 128G Memory.
    \item GPU information: Quadro RTX 6000 with 24GB of Memory.
    \item Software: CUDA 12.1, Python 3.9.21, Pytorch~\citep{Paszke19} 2.2.2, Pytorch Geometric~\citep{fey2019fast} 2.6.1.
\end{itemize}

\subsection{Robustness Evaluation Settings}
\label{robust_setups}

In our robustness study, we simulate data perturbation scenarios from three perspectives: structure, feature, and supervision signal.
Each perturbation setting is repeated 5 times with different random seeds to account for randomness, and we report the average results. Our experiments primarily focus on the node classification task.
The detailed experimental setups are as follows.

\noindent \textbf{Structure-level Robustness Evaluation Setting.} To analyze structure-level robustness, following~\citep{cai2022hypergraph}, we randomly remove or add a proportion of node–hyperedge connections (i.e., hyperlinks) in the original hypergraph and then train and evaluate HNN algorithms on the perturbed structures. The modification ratio ranges from 0 to 0.9 to simulate varying levels of noise intensity.

\noindent \textbf{Feature-level Robustness Evaluation Setting.} To study feature-level robustness, we simulate two realistic types of feature perturbations: feature noise and feature sparsity. For feature noise, following~\citep{wu2020graph}, we add independent Gaussian noise to each feature dimension of all nodes with gradually increasing amplitude. Specifically, we use the mean of the maximum feature value of each node as the reference amplitude $r$, and add Gaussian noise $\lambda \cdot r \cdot \epsilon$ to each feature dimension, where $\epsilon \sim \mathcal{N}(0,1)$ and $\lambda$ denotes the feature noise ratio. We evaluate model performance as $\lambda$ varies from 0 to 0.9 with a step size of 0.1. For feature sparsity, following~\citep{li2023gslb}, we randomly mask a certain proportion of node features by
filling them with zeros, with the sparsity ratio ranging from 0 to 0.9 at an interval of 0.1.

\noindent \textbf{Supervision-level Robustness Evaluation Setting.} We study supervision-level robustness by simulating realistic noise and sparsity scenarios. 
For label noise, following~\citep{dai2021nrgnn}, a certain proportion of training samples are randomly assigned incorrect labels by uniformly flipping them to one of the other classes.
The noise ratio varies from 0 to 0.2 in increments of 0.05. Sparsity is introduced by reducing the ratio of training nodes, with the sparsity rate ranging from 0 to 0.8 with a step size of 0.2.

\subsection{Fairness Evaluation Metrics}
\label{eval_metrics}

For fairness evaluation, we adopt two widely used group fairness metrics: demographic parity (DP)~\citep{dwork2012fairness}, and equalized odds (EO)~\citep{hardt2016equality}. 
We focus on a binary classification task, with target label $y \in \{0,1\}$ and binary sensitive attribute $s \in \{0,1\}$. 

\noindent \textbf{Demographic Parity.} If the predicted result $\hat{y}$ is independent of sensitive attributes $s$, 
i.e., $\hat{y} \perp s$, then we can consider demographic parity is achieved. 
Formally, this criterion can be expressed as:
\begin{equation}
P(\hat{y} = 1 \mid s = 0) = P(\hat{y} = 1 \mid s = 1).
\end{equation}
If a model satisfies demographic parity, the acceptance rate of different protected groups is the same.
The deviation measure $\Delta_{DP}$ in the quantitative evaluation is given by:
\begin{equation}
\Delta_{DP} = | P(\hat{y} = 1 \mid s = 0) - P(\hat{y} = 1 \mid s = 1) |,
\end{equation}
where a smaller value indicates a fairer prediction distribution across groups.

\noindent \textbf{Equalized Odds.} If the predicted outcome $\hat{y}$ and the sensitive attribute $s$ are conditionally independent 
given the ground-truth label $y$, i.e., $\hat{y} \perp s \mid y$, then we consider equalized odds is achieved. The formula for this criterion is as follows:
\begin{equation}
P(\hat{y} = 1 \mid s = 1, y = 1) = P(\hat{y} = 1 \mid s = 0, y = 1).
\end{equation}
If a model achieves equalized odds, the True Positive Rate (TPR) and False Positive Rate (FPR) 
are equal across different protected groups. The deviation measure $\Delta_{EO}$ is calculated as:
\begin{equation}
\Delta_{EO} = |P(\hat{y} = 1 \mid s = 1, y = 1) - P(\hat{y} = 1 \mid s = 0, y = 1)|,
\end{equation}
where a smaller value reflects more equitable predictive behavior across sensitive groups under the same ground-truth condition.

\section{Supplementary Experimental Results}
\label{exp:appendix}

\subsection{Experimental Results on Effectiveness Evaluation}
\label{exp:add_effect}

Table~\ref{tab:add_node_cls} shows the node classification results of all HNN algorithms on three datasets: NTU2012, ModelNet, and Ratings.

Table~\ref{tab:edge_pred},~\ref{tab:graph_cls} reports the full result of hyperedge prediction and hypergraph classification, respectively.
Tensor-based methods are not considered in the hypergraph classification task, as they lack flexibility in supporting multi-graph training.

\begin{table*}[t]
    \centering
    \small
    \caption{Additional node classification results on NTU2012, ModelNet40, and Ratings.}
    \begin{tabular}{c c c c}
        \toprule
        \textbf{Method} & \textbf{NTU2012} & \textbf{ModelNet40} & \textbf{Ratings} \\
        \midrule
        MLP & 88.59\scriptsize{$\pm$1.27}  & 96.88\scriptsize{$\pm$0.23} & 28.47\scriptsize{$\pm$0.76} \\
        CEGCN & 84.93\scriptsize{$\pm$1.12}  & 92.34\scriptsize{$\pm$0.24} & 26.65\scriptsize{$\pm$1.61} \\
        CEGAT & 84.14\scriptsize{$\pm$1.77}  & 92.02\scriptsize{$\pm$0.26} & 28.23\scriptsize{$\pm$0.50} \\
        \midrule
        HGNN & 90.13\scriptsize{$\pm$0.89}  & 97.43\scriptsize{$\pm$0.20} & 28.05\scriptsize{$\pm$0.28} \\
        HyperGCN & 75.78\scriptsize{$\pm$4.82} & 91.15\scriptsize{$\pm$3.88} & 27.34\scriptsize{$\pm$0.72} \\
        HCHA & 90.53\scriptsize{$\pm$1.00}  & 97.68\scriptsize{$\pm$0.16} & 28.33\scriptsize{$\pm$0.34} \\
        LEGCN & 89.82\scriptsize{$\pm$0.91}  & 96.82\scriptsize{$\pm$0.24} & 28.21\scriptsize{$\pm$0.50} \\
        HyperND & 88.98\scriptsize{$\pm$1.56} & 97.18\scriptsize{$\pm$0.58} & 28.32\scriptsize{$\pm$0.38} \\
        PhenomNN & 88.78\scriptsize{$\pm$0.67} & 98.28\scriptsize{$\pm$0.18} & 28.49\scriptsize{$\pm$0.41} \\
        SheafHyperGNN & 90.81\scriptsize{$\pm$0.58} & 98.30\scriptsize{$\pm$0.19} & 28.35\scriptsize{$\pm$0.57} \\
        HJRL & 88.15\scriptsize{$\pm$1.18}  & 96.33\scriptsize{$\pm$0.30} & 26.90\scriptsize{$\pm$0.55} \\
        DPHGNN & 84.77\scriptsize{$\pm$1.06} & 97.19\scriptsize{$\pm$0.17} & 28.57\scriptsize{$\pm$1.07} \\
        TF-HNN & \textbf{\textcolor{magenta}{91.69\scriptsize{$\pm$0.75}}} & \underline{98.38\scriptsize{$\pm$0.11}} & \underline{28.56\scriptsize{$\pm$0.68}} \\
        \midrule
        HNHN & 87.27\scriptsize{$\pm$1.53}  & 97.30\scriptsize{$\pm$0.27} & 27.29\scriptsize{$\pm$0.70} \\
        UniGNN & 89.86\scriptsize{$\pm$0.44} & 98.42\scriptsize{$\pm$0.08} & 28.39\scriptsize{$\pm$0.64} \\
        AllSetTransformer & 90.17\scriptsize{$\pm$1.03} & 98.07\scriptsize{$\pm$0.21} & 27.32\scriptsize{$\pm$1.11} \\
        ED-HNN & \underline{91.45\scriptsize{$\pm$0.70}} & \textbf{\textcolor{magenta}{98.51\scriptsize{$\pm$0.15}}} & 28.38\scriptsize{$\pm$0.31} \\
        HyperGT & 86.00\scriptsize{$\pm$2.05}  & 96.83\scriptsize{$\pm$0.17} & 26.58\scriptsize{$\pm$0.33} \\
        \midrule
        EHNN & 87.99\scriptsize{$\pm$0.39}  & 97.97\scriptsize{$\pm$0.17} & \textbf{\textcolor{magenta}{28.95\scriptsize{$\pm$0.81}}} \\
        T-HyperGNN & 89.15\scriptsize{$\pm$1.09} & 97.76\scriptsize{$\pm$0.34} & 24.63\scriptsize{$\pm$1.22} \\
        \bottomrule
    \end{tabular}
    \label{tab:add_node_cls}
\end{table*}

\begin{table*}[h!]
\centering
\caption{Evaluation results of hyperedge prediction.}
\label{tab:edge_pred}
\resizebox{\textwidth}{!}{\begin{tabular}{ccccccccccccc}
\toprule
\multirow{2}{*}{\textbf{Method}} & \multicolumn{2}{c}{\textbf{Cora}} & \multicolumn{2}{c}{\textbf{PubMed}} & \multicolumn{2}{c}{\textbf{Cora-CA}} & \multicolumn{2}{c}{\textbf{DBLP-CA}} & \multicolumn{2}{c}{\textbf{Actor}} & \multicolumn{2}{c}{\textbf{Pokec}} \\
\cmidrule(lr){2-3} \cmidrule(lr){4-5} \cmidrule(lr){6-7} \cmidrule(lr){8-9} \cmidrule(lr){10-11} \cmidrule(lr){12-13}
& AUROC & AP & AUROC & AP & AUROC & AP & AUROC & AP & AUROC & AP & AUROC & AP \\
\midrule
MLP          & 68.01\scriptsize{$\pm$1.23}  & 71.32\scriptsize{$\pm$1.13} & 66.00\scriptsize{$\pm$0.44} & 69.21\scriptsize{$\pm$0.61} & 71.15\scriptsize{$\pm$1.73} & 72.80\scriptsize{$\pm$1.27} & 69.19\scriptsize{$\pm$0.19} & 70.66\scriptsize{$\pm$0.36} & 54.75\scriptsize{$\pm$2.29} & 53.63\scriptsize{$\pm$1.56} & 69.69\scriptsize{$\pm$2.56} & 69.07\scriptsize{$\pm$3.03} \\
CEGCN          & 66.10\scriptsize{$\pm$2.43} & 65.50\scriptsize{$\pm$2.85} & 60.14\scriptsize{$\pm$3.97}  & 60.25\scriptsize{$\pm$3.60}  & 67.27\scriptsize{$\pm$3.78} & 71.23\scriptsize{$\pm$2.69} & 64.06\scriptsize{$\pm$1.11} & 65.07\scriptsize{$\pm$1.69} & 50.02\scriptsize{$\pm$0.01} & 50.05\scriptsize{$\pm$0.02} & 73.03\scriptsize{$\pm$2.76} & 70.08\scriptsize{$\pm$2.95} \\
CEGAT          & 72.48\scriptsize{$\pm$0.52} & 71.02\scriptsize{$\pm$0.64} & 62.20\scriptsize{$\pm$6.25} & 61.63\scriptsize{$\pm$5.99} & 69.81\scriptsize{$\pm$1.13} & 70.38\scriptsize{$\pm$1.33} & 66.50\scriptsize{$\pm$8.80} & 65.29\scriptsize{$\pm$8.21} & 56.34\scriptsize{$\pm$5.33} & 56.36\scriptsize{$\pm$5.07} & 81.01\scriptsize{$\pm$0.43} & 79.61\scriptsize{$\pm$1.60} \\
\midrule
HGNN             & 73.70\scriptsize{$\pm$1.19}  & 71.73\scriptsize{$\pm$1.57}  & 66.08\scriptsize{$\pm$9.84}  & 63.67\scriptsize{$\pm$9.02} & \underline{89.16\scriptsize{$\pm$1.11}} & \underline{89.85\scriptsize{$\pm$0.82}} & 75.44\scriptsize{$\pm$3.01} & 73.96\scriptsize{$\pm$4.91} & \textbf{\textcolor{magenta}{72.42\scriptsize{$\pm$1.96}}}  & \textbf{\textcolor{magenta}{68.79\scriptsize{$\pm$1.83}}}  & 86.09\scriptsize{$\pm$0.92} & 84.32\scriptsize{$\pm$0.95}\\
HyperGCN             & \underline{77.34\scriptsize{$\pm$1.30}}  & \underline{77.15\scriptsize{$\pm$0.33}}  & 66.46\scriptsize{$\pm$8.87}  & 64.84\scriptsize{$\pm$7.98} & \textbf{\textcolor{magenta}{92.73\scriptsize{$\pm$0.95}}} & \textbf{\textcolor{magenta}{93.42\scriptsize{$\pm$0.89}}} & \textbf{\textcolor{magenta}{89.46\scriptsize{$\pm$0.18}}} & \textbf{\textcolor{magenta}{91.39\scriptsize{$\pm$0.34}}} & 55.01\scriptsize{$\pm$8.76}  & 56.29\scriptsize{$\pm$7.44}  & \textbf{\textcolor{magenta}{91.45\scriptsize{$\pm$0.70}}} & \underline{90.76\scriptsize{$\pm$0.70}} \\
HCHA             & 73.57\scriptsize{$\pm$1.08}  & 72.24\scriptsize{$\pm$1.80}  & 63.35\scriptsize{$\pm$1.61}  & 63.13\scriptsize{$\pm$1.47} & 85.85\scriptsize{$\pm$3.27} & 84.77\scriptsize{$\pm$5.66} & 73.30\scriptsize{$\pm$3.72} & 72.09\scriptsize{$\pm$3.07} & 69.86\scriptsize{$\pm$0.95}  & \underline{66.72\scriptsize{$\pm$0.74}}  & \underline{88.81\scriptsize{$\pm$0.28}} & 88.25\scriptsize{$\pm$0.41}\\
LEGCN             & 67.16\scriptsize{$\pm$2.85}  & 68.76\scriptsize{$\pm$4.89}  & 56.39\scriptsize{$\pm$3.28}  & 54.33\scriptsize{$\pm$2.41} & 74.29\scriptsize{$\pm$0.59} & 75.95\scriptsize{$\pm$0.62} & 50.70\scriptsize{$\pm$1.40} & 50.47\scriptsize{$\pm$0.94} & 48.25\scriptsize{$\pm$3.00}  & 49.76\scriptsize{$\pm$1.29}  & 74.94\scriptsize{$\pm$1.44} & 73.89\scriptsize{$\pm$1.04}\\
HyperND            & 69.10\scriptsize{$\pm$1.28}  & 72.71\scriptsize{$\pm$1.48}  & 72.12\scriptsize{$\pm$0.78}  & 73.53\scriptsize{$\pm$0.63} & 84.01\scriptsize{$\pm$0.61} & 84.98\scriptsize{$\pm$1.07} & 78.63\scriptsize{$\pm$0.71} & 79.42\scriptsize{$\pm$0.94} & 53.12\scriptsize{$\pm$2.56}  & 52.64\scriptsize{$\pm$2.33}  & 75.77\scriptsize{$\pm$1.56} & 73.51\scriptsize{$\pm$1.62}\\
PhenomNN           & 75.71\scriptsize{$\pm$0.91}  & 75.22\scriptsize{$\pm$1.42}  & \underline{74.29\scriptsize{$\pm$0.85}}  & 72.93\scriptsize{$\pm$1.27} & 80.27\scriptsize{$\pm$1.62} & 79.59\scriptsize{$\pm$1.11} & 75.86\scriptsize{$\pm$0.86} & 75.54\scriptsize{$\pm$0.88} & 56.65\scriptsize{$\pm$3.04}  & 55.75\scriptsize{$\pm$2.87}  & 70.83\scriptsize{$\pm$2.52} & 70.17\scriptsize{$\pm$2.36} \\
SheafHyperGNN           & 70.53\scriptsize{$\pm$5.28}  & 70.93\scriptsize{$\pm$4.04}  & 68.26\scriptsize{$\pm$1.92}  & 68.07\scriptsize{$\pm$1.18} & 79.21\scriptsize{$\pm$4.53} & 75.42\scriptsize{$\pm$6.73} & 76.30\scriptsize{$\pm$1.91} & 75.41\scriptsize{$\pm$1.76} & 59.83\scriptsize{$\pm$6.77} & 59.84\scriptsize{$\pm$5.73}  & 83.44\scriptsize{$\pm$2.49} & 85.11\scriptsize{$\pm$1.80}\\
HJRL           & 58.48\scriptsize{$\pm$2.52}  & 61.02\scriptsize{$\pm$2.60}  & 59.28\scriptsize{$\pm$0.84}  & 58.63\scriptsize{$\pm$1.50} & 82.41\scriptsize{$\pm$1.90} & 85.67\scriptsize{$\pm$1.11} & OOM & OOM & 48.26\scriptsize{$\pm$0.77} & 50.00\scriptsize{$\pm$0.31}  & 84.88\scriptsize{$\pm$3.30} & 86.18\scriptsize{$\pm$2.61} \\
DPHGNN         & 66.48\scriptsize{$\pm$5.82}  & 67.23\scriptsize{$\pm$5.11}  & 60.37\scriptsize{$\pm$7.77}  & 59.86\scriptsize{$\pm$7.70} & 82.89\scriptsize{$\pm$2.28} & 83.78\scriptsize{$\pm$2.50} & OOM & OOM & 42.44\scriptsize{$\pm$5.81} & 46.60\scriptsize{$\pm$3.03} & 73.35\scriptsize{$\pm$4.59} & 73.28\scriptsize{$\pm$3.74}\\ 
TF-HNN           & 76.94\scriptsize{$\pm$0.86}  & 76.57\scriptsize{$\pm$0.71}  & 73.75\scriptsize{$\pm$0.73}  & \underline{75.54\scriptsize{$\pm$0.72}} & 74.97\scriptsize{$\pm$1.85} & 71.13\scriptsize{$\pm$1.65} & 75.70\scriptsize{$\pm$2.77} & 74.69\scriptsize{$\pm$2.26} & 54.03\scriptsize{$\pm$1.71} & 54.06\scriptsize{$\pm$1.57}  & 68.00\scriptsize{$\pm$0.97} & 67.41\scriptsize{$\pm$1.20} \\
\midrule
HNHN             & 70.13\scriptsize{$\pm$1.67}  & 68.84\scriptsize{$\pm$1.09}  & 55.67\scriptsize{$\pm$0.39}  & 53.52\scriptsize{$\pm$0.31} & 84.33\scriptsize{$\pm$1.40} & 83.49\scriptsize{$\pm$1.00} & 82.85\scriptsize{$\pm$0.70} & 82.13\scriptsize{$\pm$0.58} & \underline{69.89\scriptsize{$\pm$0.98}}  & 66.45\scriptsize{$\pm$0.74}  & 82.25\scriptsize{$\pm$1.34} & 81.72\scriptsize{$\pm$1.53}\\
UniGNN           & 73.51\scriptsize{$\pm$0.87}  & 75.23\scriptsize{$\pm$1.51}  & 74.20\scriptsize{$\pm$0.82}  & 71.76\scriptsize{$\pm$1.16} & 80.59\scriptsize{$\pm$0.98} & 82.37\scriptsize{$\pm$1.11} & 81.08\scriptsize{$\pm$0.79} & 79.39\scriptsize{$\pm$0.46} & 50.24\scriptsize{$\pm$1.26}  & 50.01\scriptsize{$\pm$0.56}  & 85.64\scriptsize{$\pm$1.20} & 84.36\scriptsize{$\pm$1.48} \\
AllSetTransformer           & 72.55\scriptsize{$\pm$2.95}  & 74.86\scriptsize{$\pm$1.85}  & 71.09\scriptsize{$\pm$2.99}  & 73.15\scriptsize{$\pm$2.49} & 76.13\scriptsize{$\pm$7.70} & 75.02\scriptsize{$\pm$8.68} & 75.12\scriptsize{$\pm$4.14} & 77.12\scriptsize{$\pm$4.22} & 55.84\scriptsize{$\pm$5.99} & 58.73\scriptsize{$\pm$4.39} & 83.65\scriptsize{$\pm$4.34} & 83.36\scriptsize{$\pm$4.72}\\
ED-HNN          & 67.24\scriptsize{$\pm$1.91}  & 69.89\scriptsize{$\pm$2.24}  & 70.09\scriptsize{$\pm$0.43}  & 72.61\scriptsize{$\pm$0.48} & 74.58\scriptsize{$\pm$1.37} & 72.94\scriptsize{$\pm$1.32} & 81.86\scriptsize{$\pm$0.67} & 84.75\scriptsize{$\pm$0.50} & 51.74\scriptsize{$\pm$2.79} & 52.27\scriptsize{$\pm$2.54}  & 85.27\scriptsize{$\pm$1.48} & 84.95\scriptsize{$\pm$1.43}\\ 
HyperGT          & 60.68\scriptsize{$\pm$4.46} & 63.02\scriptsize{$\pm$4.00} & 64.38\scriptsize{$\pm$0.58} & 67.79\scriptsize{$\pm$0.59} & 65.99\scriptsize{$\pm$2.48} & 69.66\scriptsize{$\pm$2.20} & 74.27\scriptsize{$\pm$0.24} & 72.90\scriptsize{$\pm$0.17}  & 65.18\scriptsize{$\pm$1.60} & 63.24\scriptsize{$\pm$0.53} & 81.37\scriptsize{$\pm$5.38} & 82.73\scriptsize{$\pm$5.83} \\
\midrule 
EHNN           & \textbf{\textcolor{magenta}{78.99\scriptsize{$\pm$0.99}}}  & \textbf{\textcolor{magenta}{79.54\scriptsize{$\pm$0.93}}}  & \textbf{\textcolor{magenta}{76.50\scriptsize{$\pm$0.62}}} & \textbf{\textcolor{magenta}{75.94\scriptsize{$\pm$0.70}}} & 77.83\scriptsize{$\pm$3.01}  & 78.29\scriptsize{$\pm$3.72} & \underline{87.96\scriptsize{$\pm$0.98}} & \underline{89.00\scriptsize{$\pm$0.64}} & 65.69\scriptsize{$\pm$0.46} & 65.37\scriptsize{$\pm$0.35}  & 88.63\scriptsize{$\pm$1.58} & \textbf{\textcolor{magenta}{91.31\scriptsize{$\pm$0.88}}}\\
T-HyperGNN           & 58.91\scriptsize{$\pm$1.23}  & 62.17\scriptsize{$\pm$1.58}  & 58.35\scriptsize{$\pm$4.43}  & 55.81\scriptsize{$\pm$3.71} & 66.87\scriptsize{$\pm$0.88} & 69.65\scriptsize{$\pm$0.53} & 67.17\scriptsize{$\pm$5.79} & 68.45\scriptsize{$\pm$3.85} & 49.16\scriptsize{$\pm$0.22} & 50.20\scriptsize{$\pm$0.41}  & 65.21\scriptsize{$\pm$1.21} & 66.90\scriptsize{$\pm$1.56}\\
\bottomrule
\end{tabular}}
\end{table*}

\begin{table*}[h!]
\centering
\caption{Evaluation results of hypergraph classification. Acc and F1\_ma denote the accuracy and Macro-F1, respectively. Tensor-based methods are omitted as they cannot be applied to this task.}
\label{tab:graph_cls}
\resizebox{\textwidth}{!}{\begin{tabular}{ccccccccccccc}
\toprule
\multirow{2}{*}{\textbf{Method}} & \multicolumn{2}{c}{\textbf{RHG-10}} & \multicolumn{2}{c}{\textbf{RHG-3}} & \multicolumn{2}{c}{\textbf{IMDB-Dir-Form}} & \multicolumn{2}{c}{\textbf{IMDB-Dir-Genre}} & \multicolumn{2}{c}{\textbf{Steam-Player}} & \multicolumn{2}{c}{\textbf{Twitter-Friend}} \\
\cmidrule(lr){2-3} \cmidrule(lr){4-5} \cmidrule(lr){6-7} \cmidrule(lr){8-9} \cmidrule(lr){10-11} \cmidrule(lr){12-13}
& Acc & F1\_ma & Acc & F1\_ma & Acc & F1\_ma & Acc & F1\_ma & Acc & F1\_ma & Acc & F1\_ma \\
\midrule
MLP          & 91.70\scriptsize{$\pm$1.02} & 91.43\scriptsize{$\pm$1.09} & 95.73\scriptsize{$\pm$1.86} & 95.72\scriptsize{$\pm$1.84} & 63.62\scriptsize{$\pm$1.69} & 56.98\scriptsize{$\pm$3.93} & 75.12\scriptsize{$\pm$0.70} & 71.10\scriptsize{$\pm$0.74} & 52.34\scriptsize{$\pm$0.55} & 51.60\scriptsize{$\pm$0.68} & 57.25\scriptsize{$\pm$1.81} & \underline{52.88\scriptsize{$\pm$4.57}} \\
CEGCN          & 91.50\scriptsize{$\pm$1.55} & 90.48\scriptsize{$\pm$1.42} &  98.63\scriptsize{$\pm$0.73} & 98.65\scriptsize{$\pm$0.77} & 62.66\scriptsize{$\pm$1.82}  & 55.31\scriptsize{$\pm$3.58} & 75.06\scriptsize{$\pm$0.76} & 68.98\scriptsize{$\pm$1.67} & 48.16\scriptsize{$\pm$3.87} & 47.03\scriptsize{$\pm$3.79} & 54.66\scriptsize{$\pm$5.66} & 42.16\scriptsize{$\pm$2.71} \\
CEGAT          & 88.70\scriptsize{$\pm$1.71} & 88.43\scriptsize{$\pm$1.72} & 98.80\scriptsize{$\pm$0.61} & 98.83\scriptsize{$\pm$0.59} & 63.51\scriptsize{$\pm$1.54} & 56.97\scriptsize{$\pm$4.83} & 74.12\scriptsize{$\pm$2.69} & 68.61\scriptsize{$\pm$4.73} & 49.51\scriptsize{$\pm$4.71} & 46.85\scriptsize{$\pm$4.93} & 57.32\scriptsize{$\pm$2.59} & 38.22\scriptsize{$\pm$2.54} \\
\midrule
HGNN             & 94.60\scriptsize{$\pm$1.66}   & 94.47\scriptsize{$\pm$1.84}   & 98.93\scriptsize{$\pm$0.68}  & 98.97\scriptsize{$\pm$0.65} & \underline{63.72\scriptsize{$\pm$0.62}} & \textbf{\textcolor{magenta}{57.92\scriptsize{$\pm$2.24}}} & 76.76\scriptsize{$\pm$2.66} & 72.02\scriptsize{$\pm$4.37} & 51.65\scriptsize{$\pm$2.51}  & 50.91\scriptsize{$\pm$2.92}  & 55.42\scriptsize{$\pm$2.03} & 46.81\scriptsize{$\pm$4.27}\\
HyperGCN            & 85.50\scriptsize{$\pm$1.10}  & 95.42\scriptsize{$\pm$1.09}  & 99.47\scriptsize{$\pm$0.50}  & 99.48\scriptsize{$\pm$0.49} & 62.87\scriptsize{$\pm$0.40} & 57.20\scriptsize{$\pm$2.46} & 77.53\scriptsize{$\pm$0.99} & 72.97\scriptsize{$\pm$1.08} & 51.17\scriptsize{$\pm$3.32}  & 50.48\scriptsize{$\pm$3.12}  & 56.95\scriptsize{$\pm$4.17} & 50.12\scriptsize{$\pm$5.88}\\
HCHA             & 96.60\scriptsize{$\pm$1.02}  & 96.48\scriptsize{$\pm$1.09}  & 99.33\scriptsize{$\pm$0.42}  & 99.37\scriptsize{$\pm$0.38} & 61.60\scriptsize{$\pm$2.16} & 55.37\scriptsize{$\pm$2.17} & \textbf{\textcolor{magenta}{78.12\scriptsize{$\pm$1.96}}} & \underline{73.20\scriptsize{$\pm$3.00}} & 52.43\scriptsize{$\pm$2.30}  & 51.77\scriptsize{$\pm$2.52}  & 58.17\scriptsize{$\pm$2.34} & 49.57\scriptsize{$\pm$6.84}\\
LEGCN             & 92.40\scriptsize{$\pm$1.16}  & 92.06\scriptsize{$\pm$1.19}  & 96.80\scriptsize{$\pm$0.98} & 96.78\scriptsize{$\pm$0.29} & 61.81\scriptsize{$\pm$1.32} & 56.05\scriptsize{$\pm$3.75} & 76.38\scriptsize{$\pm$1.68} & 72.03\scriptsize{$\pm$1.54}  & 53.11\scriptsize{$\pm$1.58}  & \underline{52.70\scriptsize{$\pm$1.87}}  & 56.64\scriptsize{$\pm$3.72} & \textbf{\textcolor{magenta}{53.38\scriptsize{$\pm$4.92}}}\\
HyperND            & 91.00\scriptsize{$\pm$0.95}  & 90.74\scriptsize{$\pm$1.04}  & 92.80\scriptsize{$\pm$1.95}  & 92.75\scriptsize{$\pm$1.90} & 60.74\scriptsize{$\pm$3.25} & 55.02\scriptsize{$\pm$4.95} & 75.65\scriptsize{$\pm$0.51} & 71.37\scriptsize{$\pm$1.10} & 53.88\scriptsize{$\pm$2.15}  & 49.71\scriptsize{$\pm$2.05}  & 55.27\scriptsize{$\pm$3.79} & 43.61\scriptsize{$\pm$6.51}\\
PhenomNN           & 91.10\scriptsize{$\pm$0.73}  & 90.77\scriptsize{$\pm$0.77}  & 93.47\scriptsize{$\pm$1.90}  & 93.45\scriptsize{$\pm$1.90} & 61.28\scriptsize{$\pm$1.97} & 53.71\scriptsize{$\pm$3.13} & 74.59\scriptsize{$\pm$0.61} & 70.15\scriptsize{$\pm$0.88} & 51.65\scriptsize{$\pm$3.06}  & 48.94\scriptsize{$\pm$4.55}  & 57.40\scriptsize{$\pm$3.84} & 48.26\scriptsize{$\pm$4.66} \\
SheafHyperGNN           & 96.00\scriptsize{$\pm$1.38}  & 95.96\scriptsize{$\pm$1.32}  & \underline{99.73\scriptsize{$\pm$0.33}}  & \underline{99.74\scriptsize{$\pm$0.30}} & 62.34\scriptsize{$\pm$2.06} & 56.47\scriptsize{$\pm$3.49} & 77.00\scriptsize{$\pm$1.14} & 72.78\scriptsize{$\pm$1.17} & 53.11\scriptsize{$\pm$2.39} & 52.56\scriptsize{$\pm$2.74}  & 56.49\scriptsize{$\pm$2.51} & 51.43\scriptsize{$\pm$4.42}\\
HJRL           & 96.10\scriptsize{$\pm$0.80}  & 95.98\scriptsize{$\pm$0.85}  & 99.60\scriptsize{$\pm$0.53}  & 99.57\scriptsize{$\pm$0.52} & 63.09\scriptsize{$\pm$2.83} & 56.54\scriptsize{$\pm$3.62} & \underline{77.82\scriptsize{$\pm$1.47}} & \textbf{\textcolor{magenta}{73.73\scriptsize{$\pm$1.92}}} & 51.84\scriptsize{$\pm$3.52} & 51.13\scriptsize{$\pm$3.29}  & 57.10\scriptsize{$\pm$2.79} & 44.19\scriptsize{$\pm$7.19} \\
DPHGNN         & \underline{96.80\scriptsize{$\pm$0.68}}  & \underline{96.71\scriptsize{$\pm$0.71}}  & 99.49\scriptsize{$\pm$0.65}  & 99.61\scriptsize{$\pm$0.64} & \textbf{\textcolor{magenta}{64.04\scriptsize{$\pm$2.70}}} & \underline{57.41\scriptsize{$\pm$3.96}} & 76.18\scriptsize{$\pm$1.30} & 71.59\scriptsize{$\pm$1.82} & 51.36\scriptsize{$\pm$1.72} & 49.03\scriptsize{$\pm$3.63}  & \underline{59.24\scriptsize{$\pm$2.88}} & 46.12\scriptsize{$\pm$8.49}\\ 
TF-HNN          & 95.90\scriptsize{$\pm$0.80}  & 95.88\scriptsize{$\pm$0.78}  & 98.80\scriptsize{$\pm$0.65}  & 98.84\scriptsize{$\pm$0.61} & 62.34\scriptsize{$\pm$1.76} & 55.32\scriptsize{$\pm$3.81} & 76.41\scriptsize{$\pm$1.31} & 71.89\scriptsize{$\pm$1.45} & \textbf{\textcolor{magenta}{54.85\scriptsize{$\pm$1.82}}} & \textbf{\textcolor{magenta}{52.72\scriptsize{$\pm$2.54}}}  & 56.18\scriptsize{$\pm$3.53} & 44.17\scriptsize{$\pm$8.95} \\ \midrule
HNHN             & 94.00\scriptsize{$\pm$1.90}  & 94.08\scriptsize{$\pm$1.88}  & \textbf{\textcolor{magenta}{99.92\scriptsize{$\pm$0.02}}}  & \textbf{\textcolor{magenta}{99.95\scriptsize{$\pm$0.02}}} & 62.34\scriptsize{$\pm$2.98} & 55.24\scriptsize{$\pm$3.88} & 73.65\scriptsize{$\pm$1.47} & 69.68\scriptsize{$\pm$1.18} & 52.82\scriptsize{$\pm$1.61}  & 52.68\scriptsize{$\pm$1.69}  & 58.47\scriptsize{$\pm$4.65} & 39.40\scriptsize{$\pm$3.14}\\
UniGNN           & 95.50\scriptsize{$\pm$1.38}  & 95.40\scriptsize{$\pm$1.44}  & 98.80\scriptsize{$\pm$0.27}  & 98.83\scriptsize{$\pm$0.27} & 61.06\scriptsize{$\pm$2.88} & 55.75\scriptsize{$\pm$4.01} & 77.12\scriptsize{$\pm$0.88} & 72.93\scriptsize{$\pm$1.43} & 51.46\scriptsize{$\pm$2.48}  & 48.85\scriptsize{$\pm$2.59}  & 55.88\scriptsize{$\pm$4.14} & 46.48\scriptsize{$\pm$4.90} \\
AllSetTransformer            & \textbf{\textcolor{magenta}{97.30\scriptsize{$\pm$0.98}}}  & \textbf{\textcolor{magenta}{97.26\scriptsize{$\pm$1.04}}}  & 98.80\scriptsize{$\pm$0.27}  & 98.81\scriptsize{$\pm$0.26} & 62.23\scriptsize{$\pm$1.01} & 56.26\scriptsize{$\pm$2.93} & 76.47\scriptsize{$\pm$1.38} & 72.26\scriptsize{$\pm$1.12} & 53.01\scriptsize{$\pm$2.77} & 48.21\scriptsize{$\pm$7.20}  & \textbf{\textcolor{magenta}{60.15\scriptsize{$\pm$1.70}}} & 51.52\scriptsize{$\pm$7.00}\\
ED-HNN          & 96.50\scriptsize{$\pm$0.77}  & 96.41\scriptsize{$\pm$0.78}  & 99.07\scriptsize{$\pm$0.53}  & 99.10\scriptsize{$\pm$0.51} & 62.13\scriptsize{$\pm$2.36} & 57.00\scriptsize{$\pm$4.71} & 77.12\scriptsize{$\pm$1.11} & 72.87\scriptsize{$\pm$0.44} & 52.82\scriptsize{$\pm$2.65} & 48.73\scriptsize{$\pm$2.36}  & 57.40\scriptsize{$\pm$2.66} & 42.57\scriptsize{$\pm$5.09}\\
HyperGT          & 91.60\scriptsize{$\pm$1.42} & 91.29\scriptsize{$\pm$1.53} & 96.27\scriptsize{$\pm$1.93} & 96.28\scriptsize{$\pm$1.88} & 61.49\scriptsize{$\pm$4.32} & 55.14\scriptsize{$\pm$6.02} & 73.82\scriptsize{$\pm$1.27} & 69.36\scriptsize{$\pm$1.45} & \underline{54.47\scriptsize{$\pm$1.33}} & 51.55\scriptsize{$\pm$1.08} & 54.35\scriptsize{$\pm$2.72} & 47.49\scriptsize{$\pm$5.11} \\
\bottomrule
\end{tabular}}
\end{table*}

\subsection{Experimental Results on Robustness Evaluation}
\label{exp:add_robust}

\begin{figure*}[t!]
\centering
\includegraphics[width=0.95\linewidth]{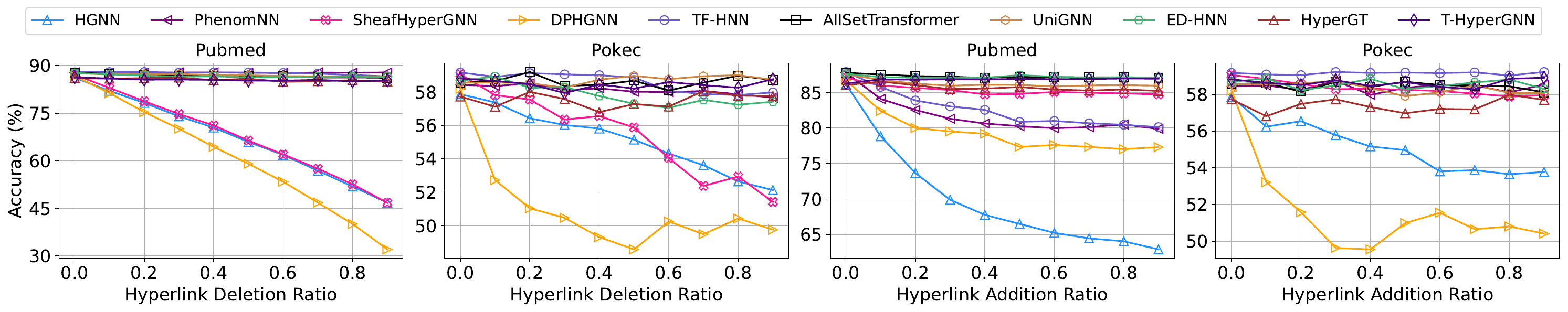}
\caption{Structure robustness analysis on Pubmed and Pokec.}
\label{fig:struc_robust_app}
\end{figure*}

\begin{figure*}[t!]
\centering
\includegraphics[width=0.95\linewidth]{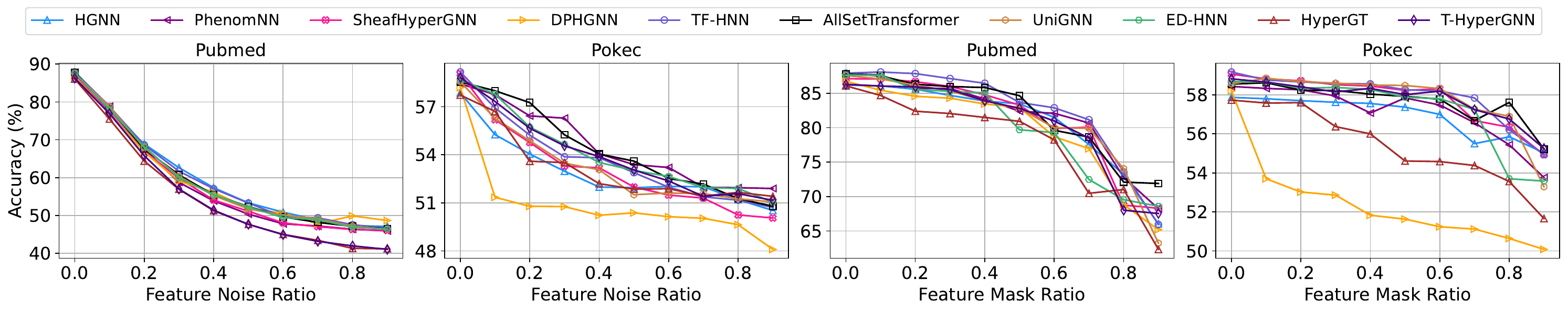}
\caption{Feature robustness analysis on Pubmed and Pokec.}
\label{fig:feat_robust_app}
\end{figure*}

\begin{figure*}[t!]
\centering
\includegraphics[width=0.95\linewidth]{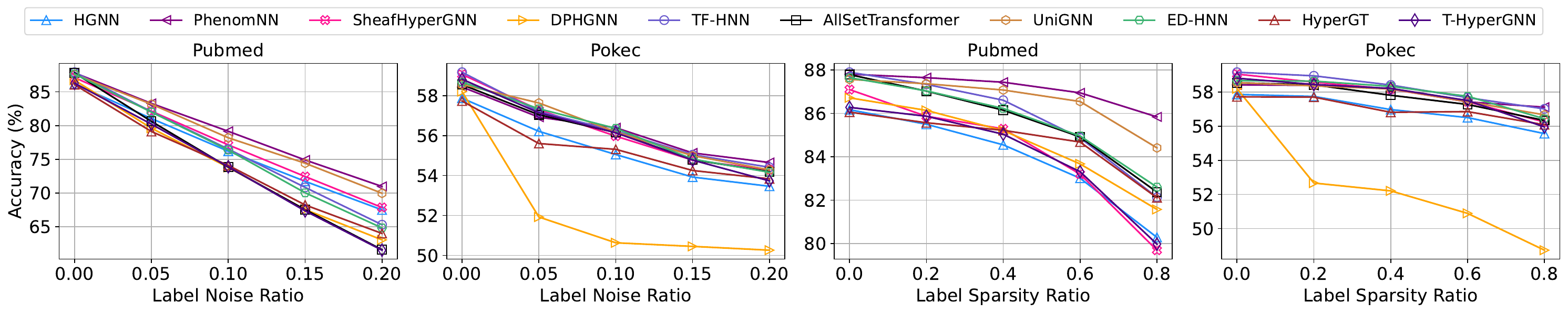}
\caption{Supervision robustness analysis on Pubmed and Pokec.}
\label{fig:super_robust_app}
\end{figure*}

Figures~\ref{fig:struc_robust_app},~\ref{fig:feat_robust_app}, and~\ref{fig:super_robust_app} show the robustness evaluation results at the structure, feature, and supervision levels on the Pubmed and Pokec datasets, respectively.
\subsection{Experimental Results on Fairness Evaluation}
\label{exp:add_fairness}

Table~\ref{tab:fairness} presents the full experimental results of fairness evaluation in terms of three metrics: accuracy (Acc), demographic parity ($\Delta_{DP}$), and equalized odds ($\Delta_{EO}$).

\begin{table*}[t]
\centering
\caption{Fairness Evaluation.}
\label{tab:fairness}
\footnotesize
\resizebox{0.9\textwidth}{!}{%
\begin{tabular}{c ccc ccc ccc}
\toprule
\multirow{2}{*}{\textbf{Method}} & \multicolumn{3}{c}{\textbf{German}} & \multicolumn{3}{c}{\textbf{Bail}} & \multicolumn{3}{c}{\textbf{Credit}} \\ 
\cmidrule(lr){2-4} \cmidrule(lr){5-7} \cmidrule(lr){8-10}
&    Acc  $\uparrow$   &     $\Delta_{DP} \downarrow$ &     $\Delta_{EO} \downarrow$   &    Acc  $\uparrow$   &     $\Delta_{DP} \downarrow$ &     $\Delta_{EO} \downarrow$  &    Acc  $\uparrow$   &   $\Delta_{DP} \downarrow$ &   $\Delta_{EO} \downarrow$   \\ \midrule
        MLP             & 67.68\scriptsize{$\pm$3.46}  &  \textbf{\textcolor{magenta}{1.78\scriptsize{$\pm$1.30}}} & 2.59\scriptsize{$\pm$0.99}  &  89.40\scriptsize{$\pm$1.76} & \underline{6.16\scriptsize{$\pm$0.93}}  & 1.79\scriptsize{$\pm$0.84}  & 79.69\scriptsize{$\pm$0.85}  &  \textbf{\textcolor{magenta}{3.43\scriptsize{$\pm$0.83}}}  & \textbf{\textcolor{magenta}{2.07\scriptsize{$\pm$0.43}}} \\ 
        CEGCN             & 69.60\scriptsize{$\pm$2.78}  & 6.19\scriptsize{$\pm$4.59} & 6.46\scriptsize{$\pm$5.48} & OOM  & OOM &  OOM & OOM  & OOM  & OOM \\
        CEGAT             & 69.12\scriptsize{$\pm$2.44}  & 9.00\scriptsize{$\pm$4.77} & 8.52\scriptsize{$\pm$3.82} &  OOM  & OOM &  OOM & OOM  & OOM  & OOM  \\
        \midrule
        HGNN             & 69.76\scriptsize{$\pm$2.50}  &  9.59\scriptsize{$\pm$3.51} & 6.90\scriptsize{$\pm$3.82} & 91.02\scriptsize{$\pm$0.54} & 7.83\scriptsize{$\pm$0.80} & 2.60\scriptsize{$\pm$1.05} & 80.21\scriptsize{$\pm$0.41}  & 5.04\scriptsize{$\pm$2.07}  & 3.46\scriptsize{$\pm$1.05} \\
        HyperGCN             & 70.40\scriptsize{$\pm$3.23}  &  6.39\scriptsize{$\pm$2.87} & 3.57\scriptsize{$\pm$1.61} & 94.72\scriptsize{$\pm$0.79} & 7.90\scriptsize{$\pm$0.95} & \underline{1.23\scriptsize{$\pm$0.52}} & 80.42\scriptsize{$\pm$0.34}  & 5.38\scriptsize{$\pm$2.61}  & 3.89\scriptsize{$\pm$1.60} \\
        HCHA            & 70.56\scriptsize{$\pm$2.51}  & 9.37\scriptsize{$\pm$3.74}  & 6.72\scriptsize{$\pm$3.05} & 91.52\scriptsize{$\pm$0.92} & 7.62\scriptsize{$\pm$0.95} & 1.40\scriptsize{$\pm$0.80} & 80.08\scriptsize{$\pm$0.43}  & \underline{3.58\scriptsize{$\pm$1.87}}  & \underline{2.47\scriptsize{$\pm$0.78}} \\
        LEGCN            & 70.88\scriptsize{$\pm$3.22}  &  7.53\scriptsize{$\pm$2.54} & 3.25\scriptsize{$\pm$2.00} &  95.02\scriptsize{$\pm$0.40} & 7.87\scriptsize{$\pm$0.62} & 1.28\scriptsize{$\pm$0.52} & \textbf{\textcolor{magenta}{80.48\scriptsize{$\pm$0.37}}}  & 4.31\scriptsize{$\pm$2.24}  & 3.15\scriptsize{$\pm$1.12} \\
        HyperND            & 71.04\scriptsize{$\pm$2.61}  & 7.37\scriptsize{$\pm$4.70} & 3.67\scriptsize{$\pm$3.10} & 89.75\scriptsize{$\pm$2.41} & 7.92\scriptsize{$\pm$1.52} & 3.19\scriptsize{$\pm$2.22} & 80.02\scriptsize{$\pm$0.49}  & 4.14\scriptsize{$\pm$2.22}  &  2.50\scriptsize{$\pm$0.72} \\
        PhenomNN           & 70.96\scriptsize{$\pm$2.85}  & 3.54\scriptsize{$\pm$3.07} & 1.60\scriptsize{$\pm$1.94} & 91.71\scriptsize{$\pm$1.13} & 10.83\scriptsize{$\pm$1.64} & 1.94\scriptsize{$\pm$0.40} &  OOM  &  OOM   &  OOM \\
        SheafHyperGNN           & 70.64\scriptsize{$\pm$3.29}  & 8.14\scriptsize{$\pm$3.25} & 5.23\scriptsize{$\pm$2.05} & OOM &  OOM  &  OOM  &  OOM  &  OOM   &  OOM  \\
        HJRL           & 69.92\scriptsize{$\pm$3.46}  &  3.52\scriptsize{$\pm$2.70} & 3.05\scriptsize{$\pm$1.82} & OOM & OOM  & OOM & OOM & OOM  & OOM \\ 
        DPHGNN          & 70.24\scriptsize{$\pm$3.25}  & \underline{2.25\scriptsize{$\pm$0.49}}  & \textbf{\textcolor{magenta}{1.38\scriptsize{$\pm$0.77}}}  & 93.41\scriptsize{$\pm$0.93} & 8.07\scriptsize{$\pm$1.20} & 2.05\scriptsize{$\pm$1.22} & OOM & OOM  & OOM \\
        TF-HNN          & 70.48\scriptsize{$\pm$3.14}  &  5.27\scriptsize{$\pm$3.09} & 4.19\scriptsize{$\pm$2.23} & \underline{95.33\scriptsize{$\pm$0.25}} & 7.96\scriptsize{$\pm$0.65} & \textbf{\textcolor{magenta}{1.03\scriptsize{$\pm$0.67}}} & \underline{80.46\scriptsize{$\pm$0.36}} & 4.93\scriptsize{$\pm$2.44}  & 3.43\scriptsize{$\pm$1.43} \\ 
        \midrule
        HNHN           & 69.52\scriptsize{$\pm$3.62}  & 4.01\scriptsize{$\pm$2.76}  &  \underline{1.59\scriptsize{$\pm$1.60}} & 90.76\scriptsize{$\pm$1.30} & \textbf{\textcolor{magenta}{6.03\scriptsize{$\pm$1.43}}} & 3.04\scriptsize{$\pm$1.34} & 78.00\scriptsize{$\pm$0.23} & 5.70\scriptsize{$\pm$3.11}  & 4.67\scriptsize{$\pm$3.10} \\
        UniGNN           & \underline{71.07\scriptsize{$\pm$2.70}}  &  5.08\scriptsize{$\pm$3.03} & 2.80\scriptsize{$\pm$1.32} & 91.30\scriptsize{$\pm$1.47} & 9.42\scriptsize{$\pm$1.68} & 3.94\scriptsize{$\pm$2.18} & 80.44\scriptsize{$\pm$0.37} & 3.90\scriptsize{$\pm$2.40}  & 2.85\scriptsize{$\pm$1.33} \\
        AllSetTransformer          & 70.48\scriptsize{$\pm$3.11}  &  4.47\scriptsize{$\pm$3.39} & 3.50\scriptsize{$\pm$3.38} & \textbf{\textcolor{magenta}{96.26\scriptsize{$\pm$1.83}}} & 8.36\scriptsize{$\pm$0.85} & 1.95\scriptsize{$\pm$1.10} & 80.40\scriptsize{$\pm$0.44} &  4.46\scriptsize{$\pm$2.96}  & 3.44\scriptsize{$\pm$1.60} \\
        ED-HNN           & 70.16\scriptsize{$\pm$3.15}  & 4.06\scriptsize{$\pm$3.05} & 4.07\scriptsize{$\pm$2.75} & 94.26\scriptsize{$\pm$0.77} & 8.05\scriptsize{$\pm$0.64} & 1.51\scriptsize{$\pm$0.26} & OOM & OOM  & OOM \\ 
        HyperGT             & 68.88\scriptsize{$\pm$2.01} & 5.05\scriptsize{$\pm$2.88} & 4.36\scriptsize{$\pm$2.59} & 94.33\scriptsize{$\pm$0.62} & 7.68\scriptsize{$\pm$1.13} & 1.64\scriptsize{$\pm$1.37} & 79.83\scriptsize{$\pm$0.39}  & 4.17\scriptsize{$\pm$2.50}  & 2.69\scriptsize{$\pm$1.97} \\
        \midrule
        EHNN           & 70.40\scriptsize{$\pm$3.07}  & 2.87\scriptsize{$\pm$5.73}  & 2.34\scriptsize{$\pm$4.69}  & 93.62\scriptsize{$\pm$1.75} & 9.29\scriptsize{$\pm$1.60}  & 2.88\scriptsize{$\pm$1.23} & 80.34\scriptsize{$\pm$0.47} & 4.51\scriptsize{$\pm$2.77}  & 3.13\scriptsize{$\pm$1.75} \\
        T-HyperGNN           & \textbf{\textcolor{magenta}{71.20\scriptsize{$\pm$1.82}}}  &  8.99\scriptsize{$\pm$6.52} & 6.80\scriptsize{$\pm$5.02} &  OOM & OOM  & OOM & OOM & OOM  & OOM \\
\bottomrule
 
\end{tabular}}
\end{table*}

\section{Further Discussions}

\subsection{Related Works}

Hypergraph neural networks (HNNs)~\citep{yadati2019hypergcn,prokopchik2022nonlinear,wang2023equivariant,xie2025k} have been promising tools for handling learning tasks involving higher-order data, with notable applications in various fields, such as social network analysis~\citep{sun2023self}, bioinformatics~\citep{li2025dghnn}, and recommender systems~\citep{li2025deep}. 
However, there exists no established benchmark specifically dedicated to comprehensively evaluating hypergraph neural networks. 
In this section, we introduce a broader range of related studies concerning the comparative evaluations of HNNs, providing sufficient context for our benchmark work.

Kim et al.~\citep{kim2024survey} recently presented the
first survey dedicated to HNNs, with an in-depth and step-by-step
guide. The survey comprehensively reviews existing HNN architectures, training strategies, and applications, establishing a 
foundational understanding crucial for advancing the field of HNNs. To further understand the expressive power of HNNs, Wang et al.~\citep{wang2025generalization} conduct the first theoretical analysis on the generalization performance of 
distinct HNN architectures, offering practical guidance for improving HNNs’ effectiveness. Nevertheless, systematic empirical evaluations of different HNN algorithms remain scarce, leaving a limited understanding of their comparative performance in practice. 
To facilitate the reproducibility and empirical evaluation of HNN algorithms, several open-sourced libraries have been developed in recent years.
HyFER~\citep{hwang2021hyfer} is a well-modularized framework for implementing and evaluating HNNs, dividing
the entire learning process into data, model, and task components.
Moreover, to address the scalability problem that most existing implementations
suffer from, HyFER is built on top of Deep Graph Library (DGL)~\citep{wang2019deep},
which is a highly efficient open-sourced library for GNNs. DHG~\citep{gao2022hgnn+} is an open-sourced PyTorch-based toolbox designed for general HNNs. 
It supports various hypergraph
preprocessing methods (e.g., sampling, expansion) and convolution operators, facilitating the evaluation of HNNs. 
TopoX~\citep{hajij2024topox} is a suite of Python
packages for machine learning on topological domains. These
packages enhance and generalize functionalities found in mainstream hypergraph computations and learning tools, enabling them on topological domains.
TopoBench~\citep{telyatnikov2024topobench} is a modular Python library that standardizes benchmarking and accelerates research in Topological Deep Learning (TDL). It supports training and comparing Topological Neural Networks (TNNs) across diverse domains, including graphs, simplicial complexes, cellular complexes, and hypergraphs.
However, these libraries do not fully cover the latest HNN algorithms, datasets, and evaluation tasks, and they provide only limited empirical results without offering an in-depth and comprehensive analysis of existing HNN methods.

To fill the gap, we develop DHG-Bench, the first comprehensive benchmark tailored explicitly for HNNs. Distinguished by its broad coverage, DHG-Bench spans a wide range of algorithms, datasets, and evaluation tasks, thereby establishing a standardized and versatile framework for deep hypergraph learning research.
Moreover, it provides comprehensive and systematic empirical evaluations that uncover the strengths and limitations of different algorithms. By offering such in-depth quantitative analyses, our benchmark fosters deeper insights into the challenges and opportunities of HNNs, thereby advancing the state-of-the-art in this emerging field.

\begin{figure*}[t]
\centering
\includegraphics[width=\textwidth]{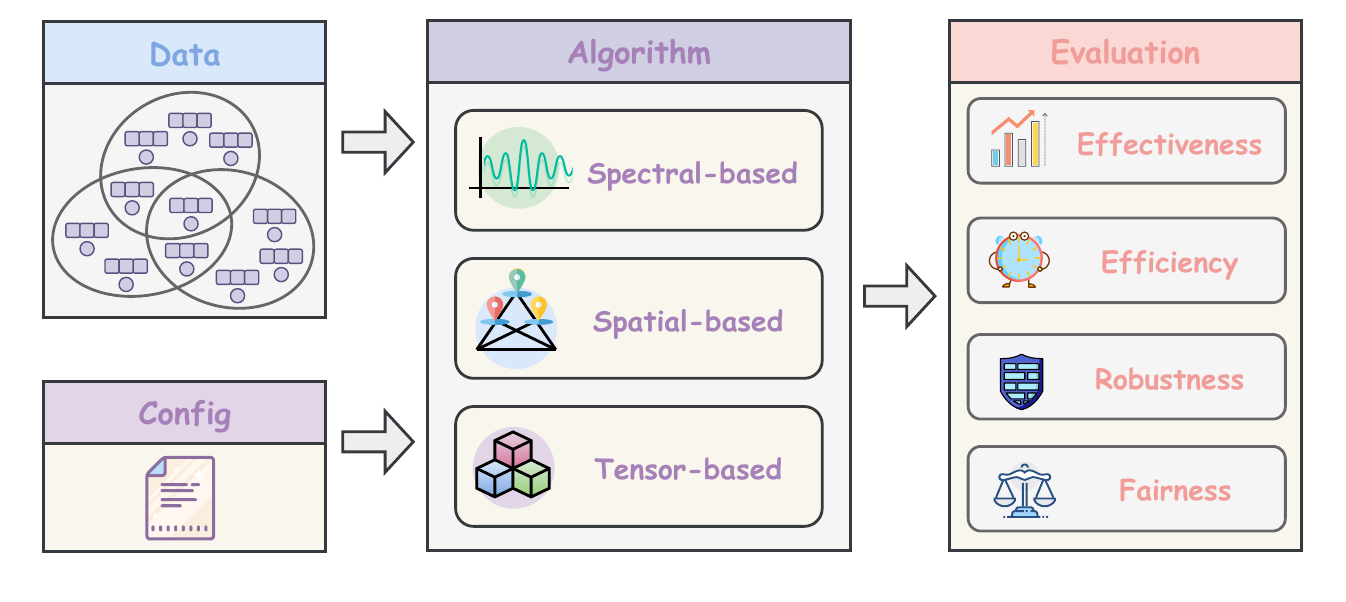} 
\caption{The package structure of DHG-Bench, which mainly consists of four modules.}
\label{fig:package}
\end{figure*}

\subsection{Future Directions}
\label{future_direct}

Drawing upon our empirical analyses in the main text, we point out some promising future directions for the deep hypergraph learning community.
\begin{itemize}[leftmargin=15pt]
  \item \textbf{Developing adaptive HNN methods for diverse datasets and tasks.} 
  Our experiments in Section~\ref{exp:effectiveness} reveal that existing HNN architectures show substantial performance disparities across datasets and tasks, limiting their applicability in diverse scenarios. 
  Future research should focus on designing adaptive HNN architectures and training techniques that can better accommodate the unique characteristics of datasets from different domains and varying task granularities, thereby enhancing the generalization ability of HNNs.
  \item \textbf{Improving the efficiency of HNN methods.} Observations in Section~\ref{exp:efficiency} indicate that many advanced HNN methods fail to balance efficiency and predictive performance, and often run out of memory on large-scale datasets. As the size of hypergraphs continues to grow
exponentially, a key area of future research is the reduction of memory and computational complexity in HNN algorithms while maintaining satisfactory model utility. 
  Inspired by the favorable efficiency–effectiveness trade-off achieved by TF-HNN, it would be promising to devise more powerful decoupled architectures specifically tailored for HNN.
  \item \textbf{Developing more robust HNN methods.} Our experimental results in Section~\ref{exp:robustness} show that HNN algorithms are affected by different types of data perturbations and are particularly vulnerable to those at the feature and supervision levels. Future work should emphasize enhancing the robustness of HNNs to resist varying degrees of data noise and even adversarial attacks, thereby ensuring reliable performance in a wide range of industrial applications.
  \item \textbf{Developing fairness-aware HNN methods.} 
  Empirical evidence in Section~\ref{exp:fairness} suggests that HNNs are more prone to biased predictions than traditional MLPs. Future research should investigate the theoretical mechanisms through which high-order message passing exacerbates fairness issues and then develop fairness-aware HNN methods that mitigate such discriminatory behavior. Progress in this direction is essential to ensure the safe adoption of HNNs in high-stakes real-world applications such as crime prediction and credit evaluation.
\end{itemize}

\section{Package}

We have developed DHG-Bench~\footnote{\url{https://github.com/Coco-Hut/DHG-Bench}}, an open-sourced package that provides a comprehensive and unbiased platform for evaluating HNN algorithms and supporting future research in this domain.

As shown in Figure~\ref{fig:package}, the code structure is well-designed to ensure fair experimental setups across different algorithms,
easy reproduction of the experimental results, and support for flexible assembly of models for experiments. The DHG-Bench
consists of the following four key modules. 
\circled{1} The Config module
includes the files that define the necessary hyperparameters and settings. 
\circled{2} The Data module is used to load and preprocess datasets. 
\circled{3} The Algorithm module has 17 built-in state-of-the-art algorithms, covering three representative categories: spectral-based, spatial-based, and tensor-based methods.
\circled{4} The evaluation module supports multi-faceted testing of algorithmic performance, encompassing effectiveness, efficiency, robustness, and fairness.

\end{document}

%% file: math_commands.tex
\usepackage{amsmath,amsfonts,bm}

\def\eqref#1{equation~\ref{#1}}
\def\1{\bm{1}}

\DeclareMathAlphabet{\mathsfit}{\encodingdefault}{\sfdefault}{m}{sl}
\SetMathAlphabet{\mathsfit}{bold}{\encodingdefault}{\sfdefault}{bx}{n}